\documentclass[10pt,twocolumn,letterpaper]{article}

\usepackage{iccv}
\usepackage{times}
\usepackage{epsfig}
\usepackage{graphicx}
\usepackage{amsmath}
\usepackage{amssymb}

\usepackage{subfigure}
\usepackage{arydshln}

\usepackage[table,xcdraw]{xcolor}
\usepackage{multirow}
\usepackage{diagbox}
\usepackage{booktabs}
\usepackage[normalem]{ulem}
\useunder{\uline}{\ul}{}

\usepackage[breaklinks=true,bookmarks=false]{hyperref}

\iccvfinalcopy 

\def\iccvPaperID{****} 

\ificcvfinal\fi

\begin{document}

\title{Learning Residue-Aware Correlation Filters and Refining Scale Estimates with the GrabCut for Real-Time UAV Tracking}

\author{Shuiwang Li\quad Yuting Liu\quad Qijun Zhao\quad Ziliang Feng\\
College of Computer Science, Sichuan University\\
{\tt\small lishuiwang0721@163.com,yuting.liu@stu.scu.edu.cn,\{qjzhao,fengziliang\}@scu.edu.cn}
}

\maketitle
\ificcvfinal\thispagestyle{empty}\fi

\begin{abstract}

Unmanned aerial vehicle (UAV)-based tracking is attracting increasing attention and developing rapidly in applications such as agriculture, aviation, navigation, transportation and public security. Recently, discriminative correlation filters (DCF)-based trackers have stood out in UAV tracking community for their high efficiency and appealing robustness on a single CPU. However, due to limited onboard computation resources and other challenges the efficiency and accuracy of existing DCF-based approaches is still not satisfying. In this paper, we explore using segmentation by the GrabCut to improve the wildly adopted discriminative scale estimation in DCF-based trackers, which, as a mater of fact, greatly impacts the precision and accuracy of the trackers since accumulated scale error degrades the appearance model as online updating goes on. Meanwhile, inspired by residue representation, we exploit the residue nature inherent to videos and propose residue-aware correlation filters that show better convergence properties in filter learning. Extensive experiments are conducted on four UAV benchmarks, namely, UAV123@10fps, DTB70, UAVDT and Vistrone2018 (VisDrone2018-test-dev). The results show that our method achieves state-of-the-art performance.

\end{abstract}

\section{Introduction}

Unmanned aerial vehicle (UAV)-based tracking is an emerging task and has attracted increasing attention in recent years. It has various applications, e.g., aerial patrolling\cite{karaduman2019uav}, disaster response \cite{2017Aerial}, autonomously landing \cite{2016Monocular}, wildlife protection \cite{2015Towards}. Compared with general tracking scenes, UAV tracking faces more onerous challenges, e.g., motion blur, severe occlusion, extreme visual angle and scale change, and scarce computation resources \cite{2019Correlation}.
\begin{figure}[h]
	\centering
	\includegraphics[width=0.475\textwidth,height=0.24\textwidth]{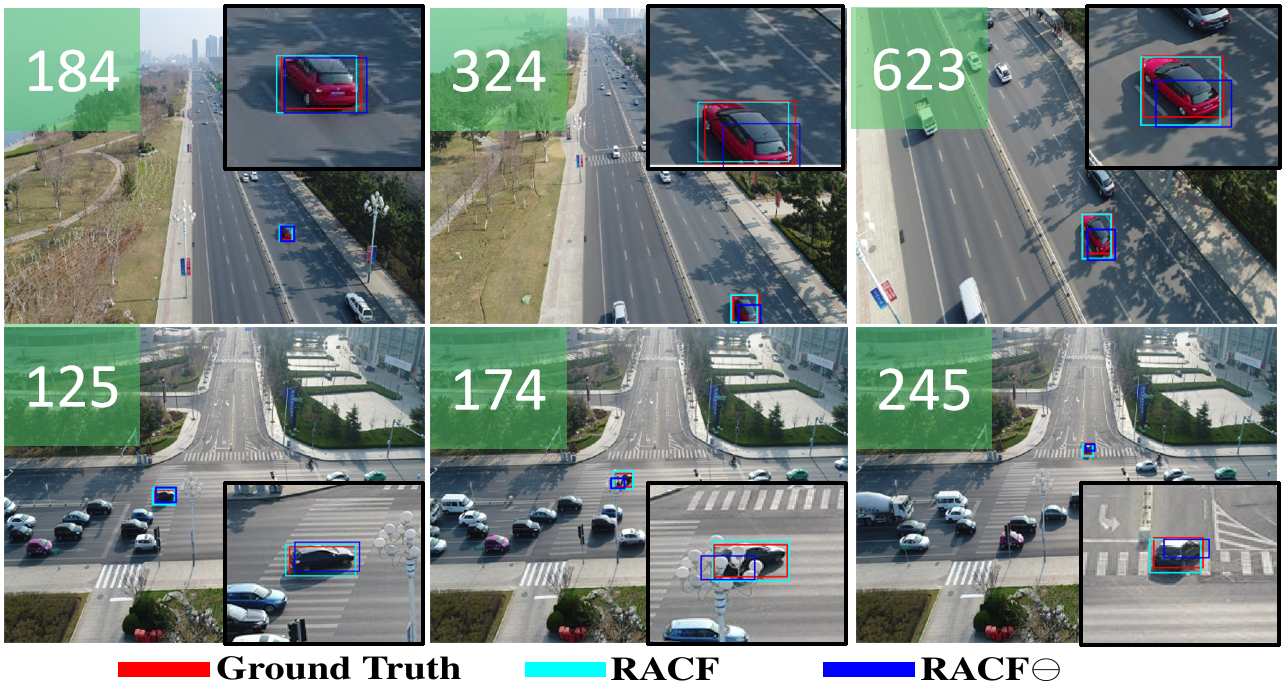}
	\caption{Comparison of our trackers RACF and RACF$\ominus$. RACF uses the GrabCut to refine the scale estimates whereas RACF$\ominus$ uses the existing discriminative scale estimation only.} \label{fig:visual_examples0}
\end{figure}
Despite deep learning-based approaches have achieved great success and are promising in dealing with these challenges \cite{danelljan2020probabilistic,lukezic2020d3s,voigtlaender2020siam,bhat2020know,li2019gradnet,wang2019fast}, its efficiency is unsatisfactory in the limitation of power capacity and computational resources onboard UAVs. In contrast, discriminative correlation filters (DCF)-based methods are more favorable in efficiency for UAV tracking thanks to the fast Fourier transform (FFT) \cite{huang2019learning,li2020autotrack, 2020Object,li2020keyfilter,li2020training,lin2020bicf,li2019augmented,he2020towards} but they are not comparable with deep learning-based method in terms of precision and accuracy. One important reason is the failing to recognize the problems with the discriminative scale estimation \cite{li2014a,danelljan2017discriminative} used in DCF-based trackers. 
First, applying the filter at limited predefined multiple resolutions is inaccurate in estimating continuous changes of scale. Second, a fixed aspect ratio of target size is usually adopted to balance accuracy and speed in scale estimation, which is very rough and degrades the tracking greatly in situations where targets undergo extreme visual angle and appearance variations. See Fig. \ref{fig:visual_examples0} for examples. As can be seen, the RACF$\ominus$ tracker using the existing discriminative scale estimation performs badly when the two target cars undergo big visual angle changes. 
Therefore, the existing discriminative scale estimation is inaccurate, which in fact greatly impacts the precision and accuracy of the trackers, since accumulated scale error degrades the appearance model as online updating goes on.  
In this paper, we study using the segmentation algorithm GrabCut \cite{2004GrabCut} to improve scale estimation of DCF-based trackers. The method is demonstrated by the tracker RACF shown in Fig. \ref{fig:visual_examples0} .

On the other hand, most DCF-based methods improve precision and accuracy at the cost of efficiency. For instance, the aberrance repressed correlation filters (ARCF) \cite{huang2019learning} has achieved state-of-the-art performance in UAV tracking by repressing aberrances happening during the tracking process, but it takes five iterations of ADMM (Alternating Direction Method of Multipliers) at each frame for model update, which is time-consuming compared with two iterations in its baseline the background-aware correlation filters
(BACF) \cite{2017Learning}. Inspired by the fast and stable optimization of Residual Networks (ResNets) \cite{he2016deep}, we exploit the residue nature inherent to videos and propose residue-aware correlation filters (RACF) that show better convergence properties in filter learning, which, to the best of our knowledge, has not been studied in DCF-based trackers before.
In addition, ARCF does not use spatial and temporal regularizations which have been proved effective for suppressing boundary effects and avoiding overfitting to inaccurate and noisy samples \cite{danelljan2015learning,li2018learning,li2020asymmetric,li2020autotrack}. In this paper, we succeed to add spatial and temporal regularizations to boot performance with little additional computational cost. 
The contributions of this paper are summarized as follows: 
\begin{itemize}
	\item We propose a novel scale estimation approach for DCF-based trackers by using the GrabCut \cite{2004GrabCut} algorithm to refine the discriminative scale estimates, which can be incorporated easily into any tracking method with discriminative scale estimation to improve precision and accuracy.
	\item We propose a novel regularization to model the residue between two neighboring frames, resulting in what we call residue-aware correlation filters, which show better convergence properties in filter learning. Meanwhile, we add spatial and temporal regularizations to boot performance with little additional computational cost. 
	\item We demonstrate the proposed methods on four UAV benchmarks, namely, UAV123@10fps \cite{2016A}, DTB70 \cite{li2017visual}, UAVDT \cite{du2018the} and Vistrone2018 (VisDrone2018-test-dev) \cite{wen2018visdrone}. Experimental results show that the proposed approaches achieves state-of-the-art performance.
\end{itemize}


\section{Related Works}
\subsection{Discriminative correlation filters}
Using DCF for visual tracking starts with the minimum output sum of squared error (MOSSE) filter \cite{bolme2010visual}. Afterwards, kernel trick \cite{2015High}, discriminative scale estimation \cite{li2014a,danelljan2017discriminative}, spatial and temporal regularizations \cite{danelljan2015learning,li2018learning}, deep features \cite{ma2015hierarchical,2016Convolutional}, attention \cite{he2020towards}, context and background information \cite{2017Context,2017Learning} and so on were introduced for improvement. Among these DCF-based trackers, we are interested in the background-aware correlation filters (BACF) \cite{2017Learning}. For one thing, it expands search region with lower computational cost and also has some theoretical advantages as shown in \cite{li2020asymmetric}, an equivalent approach but formulated and optimized differently. For another, BACF equipped with an aberrance regularization and an enhanced feature extractor, i.e., the aberrance repressed correlation filters (ARCF) \cite{huang2019learning}, has achieved state-of-the-art performance in multiple benchmarks specified for UAV tracking. These two methods are in close connection with our work here.

\subsection{Residual representations}
In visual tracking, residual learning was applied to capture the difference between the base layer output and the ground truth to reduce model degradation during online update in \cite{song2017crest}. Before that residual representations had found their usefulness in image retrieval, recognition and restoration \cite{he2016deep,vedaldi2010vlfeat,chatfield2011the}, video compression \cite{tsai2018learning} and equation solving \cite{nash2000multigrid}. It has been shown
\cite{szeliski2006locally,briggs2000multigrid} that the solvers aware of the residual nature of the solutions converge much faster than standard solvers \cite{szeliski2006locally,nash2000multigrid,he2016deep}. To understand the optimization landscape of the well-known ResNets \cite{he2016deep}, it has been proved that residual networks eliminate singularities and reduce shattered gradient, leading to numerical stability and easier optimization \cite{orhan2018skip,balduzzi2017the,kawaguchi2016deep}. In the literature, residuals can be the difference between the target and the model output, the difference between the input and the output, the difference between a sample and the sample mean.
In this paper, a residue is formed by subtracting the feature of the current frame from that of the previous one and the residue representation of the current frame is the sum of the residue and the feature of the previous frame. Since, approximately, videos are temporally continuous, the residues are usually of small values and entropy, which can be used to speed up the learning of correlation filters.

\subsection{Scale estimation}
There are three typical discriminative scale estimation strategies in DCF-based visual tracking \cite{2021Efficient,2017A}. i) Multi-resolution translation filter (MRTF) defines a scale pool and estimates target scale by applying the translation filter on images of different scales \cite{li2014a,2016Target}. ii) Joint scale-space filter
(JSSF) jointly estimates the translation and scale of the 
target by simultaneously maximizing the
scores of translation and scale filters with a three-dimensional Gaussian function as the desired response. iii) Separate scale filter (SSF) learns a separate 1-dimensional scale correlation filter for scale estimation independent of translation \cite{2014Accurate,danelljan2017discriminative}. The SSF strategy has been popular in DCF-based trackers for its effectiveness, efficiency, robustness and easy integration. Although there are some variants such as using subgrid interpolation for efficiency \cite{danelljan2017discriminative}, combining a points tracker for abrupt scale changes \cite{2019Patch}, defining rotation-aware correlation filters \cite{2019Rotation} and utilizing  deep features for scale filter \cite{2021Efficient}, these methods for scale estimation essentially remain detection-based and therefore inaccurate. 

Recently, many researchers have noticed the close relation between video object segmentation and visual tracking and combined them to improve respective performance \cite{wang2019fast,lukezic2020d3s,chen2020state,sun2020fast}. It makes sense since accurate segmentation results provide reliable object observations for tracking and in turn precise target states provided by a reliable tracker correctly guide and speed up object segmentation. Unfortunately, they are deep learning-based approaches, not suitable for UAV tracking with limited onboard resources.
There do exist plenty of CPU-based methods earlier by combing tracking and segmentation to overcome the inaccurate object description with rectangular bounding box, especially for tracking non-rigid objects \cite{aeschliman2010a,ren2007tracking,duffner2013pixeltrack,godec2011hough}. However, both appearance model and scale estimation in these approaches are strongly dependent on traditional segmentation algorithms which may be quite accurate in certain circumstances but are not robust generally. In this paper, we leverage the efficiency and robustness of SSF and the accuracy of segmentation by the GrabCut \cite{2004GrabCut} for scale estimation in UAV tracking by taking the former as a crude estimate and the latter as a potential refinement, which, to the best of our knowledge, has not been studied before.

\section{Revisit BACF and ARCF}

Given the vectorized samples of $D$ channels $\{\mathbf{x}^d_k |\mathbf{x}^d_k\in R^N\}_{d=1}^{D}$ at frame $k$ and the vectorized ideal response $\mathbf{y} \in R^N$, BACF aims to minimize the following objective:
\begin{equation}
	E(\mathbf{f}_k)=\frac{1}{2}||\mathbf{y}-\sum_{d=1}^{D}\mathbf{x}^d_k\ast \mathbf{P}\mathbf{f}^d_k||_2^2 + \frac{\lambda}{2}\sum_{d=1}^{D}||\mathbf{f}^d_k||_2^2 ,
\end{equation}
where $\mathbf{f}^d_k\in R^M$ is the filter to be learned and $\mathbf{P} \in R^{N \times M}$ is a padding matrix which symmetrically pads $\mathbf{f}^d_k$ with zeros such that $\mathbf{P}\mathbf{f}^d_k \in R^N$ and $\mathbf{P}^\textup{T}\mathbf{P}\mathbf{f}^d_k=\mathbf{f}^d_k$. $\ast$ denotes the correlation operator, the operator $^\textup{T}$ transposes a matrix and $M \ll N$. $\lambda$ is a penalty coefficient. ARCF introduces an additional regularization and proposes to minimize the following objective:
\begin{equation}\label{EQ_ARCF}
	\begin{aligned}
		E(\mathbf{f}_k)=\frac{1}{2}||\mathbf{y}-\sum_{d=1}^{D}\mathbf{x}_k^d\ast \mathbf{P}\mathbf{f}_k^d||_2^2+\frac{\lambda }{2}\sum_{d=1}^{D}||\mathbf{f}_k^d||_2^2\\
		+\frac{\gamma  }{2}||\mathbf{M}_{k-1}[{\psi}_{p,q} ]-\mathbf{M}_k||_2^2 ,
	\end{aligned}
\end{equation}
where $\mathbf{M}_k=\sum_{d=1}^{D}\mathbf{x}_k^d\ast \mathbf{P}\mathbf{f}_k^d$ denotes the response map at frame $k$, $(p,q)$ is the two-dimensional location offset of two peaks in $\mathbf{M}_k$ and $\mathbf{M}_{k-1}$, and $[{\psi}_{p,q}]$ indicates the shifting operation with which the two peaks coincide. $\gamma$ is the penalty coefficient.

\section{Proposed Approach}
\subsection{Residue-aware correlation filters}\label{subsection_RRCF}
The proposed residue-aware correlation filter with spatial and temporal regularizations is as follows:
\begin{equation}\label{EQ_RRCF}
	\begin{split}
		E(\mathbf{f}_k)=\frac{1}{2}||\mathbf{y}-\sum_{d=1}^{D}\mathbf{x}_k^d\ast \mathbf{P}\mathbf{f}_k^d||_2^2+\frac{\eta }{2}||\sum_{d=1}^{D}{\boldsymbol{\delta}} _k^d\ast \mathbf{P}\mathbf{f}_k^d||_2^2\\
		+\frac{\theta }{2}\sum_{d=1}^{D}||\mathbf{w}\odot  \mathbf{f}_k^d||_2^2+\frac{\tau }{2}\sum_{d=1}^{D}|| \mathbf{f}_k^d-\mathbf{f}_{k-1}^d||_2^2+\frac{\lambda }{2}\sum_{d=1}^{D}||\mathbf{f}_k^d||_2^2,
	\end{split}
\end{equation}
where ${\boldsymbol{\delta}} _k^d=\mathbf{x}_k^d-\mathbf{x}_{k-1}^d$ denotes sample residue at the $k$th frame, $\mathbf{w}$ is a vectorized spatial regularization weight of bowl shape and $\odot $ denotes the Hadamard product. Parameters $\eta$, $\theta$ and $\tau$ are predefined penalty coefficients for residue-aware, spatial and temporal regularizations respectively. Eq. (\ref{EQ_RRCF}) will be transformed into the frequency domain and optimized with ADMM. Notice that the aberrance term of ARCF, i.e., $||M_{k-1}[{\psi}_{p,q} ]-M_k||_2^2$ in Eq. (\ref{EQ_ARCF}), reduces to the proposed residue one, i.e., $||{\boldsymbol{\delta}} _k^d\ast \mathbf{P}\mathbf{f}_k^d||_2^2$ in Eq. (\ref{EQ_RRCF}), when ${\psi}_{p,q}\equiv{\psi}_{0,0}$, namely no shifting operation is required, and $\mathbf{f}_{k}^d=\mathbf{f}_{k-1}^d$. As a reward of our formulation, the residue nature manifests itself and it takes only two iterations of ADMM at each frame in our algorithm, a considerable computational saving compared with five in ARCF.
\subsubsection{Transformation into frequency domain}
For efficiency, correlation filters are typically solved in the frequency domain \cite{danelljan2015learning,2017Learning,huang2019learning}. Let $\mathbf{F}$ denote the Fourier transform such that $\mathbf{F^{-1}}=\mathbf{F}^\textup{H}$, where the operator $^\textup{H}$ computes the conjugate transpose on a complex vector or matrix, and $\hat{\mathbf{z}}=\mathbf{Fz}$ be the Fourier transform of $\mathbf{z}$. Equipped with an auxiliary variable $\hat{\mathbf{g}}_k$, Eq. (\ref{EQ_RRCF}) can be expressed in the frequency domain as:
\begin{equation}\label{EQ_RRCF_FT}
	\begin{split}
		E(\mathbf{f}_k,\hat{\mathbf{g}}_k)=\frac{1}{2}||\mathbf{\hat{y}}-\mathbf{X}_k \hat{\mathbf{g}}_k||_2^2+\frac{\eta }{2}||\mathbf{\Delta} _k^d \hat{\mathbf{g}}_k||_2^2\\
		+\frac{\theta }{2}||\mathbf{W}  \mathbf{f}_k||_2^2+\frac{\tau }{2}|| \mathbf{f}_k-\mathbf{f}_{k-1}||_2^2+\frac{\lambda }{2}||\mathbf{f}_k||_2^2,\\
		s.t. \qquad \hat{\mathbf{g}}_k=(\mathbf{I}_D \otimes \mathbf{FP})\mathbf{f}_k
	\end{split}
\end{equation}
where $\mathbf{X}_k=[\textup{diag}(\mathbf{\hat{x}}_k^1)^\textup{H},...,\textup{diag}(\mathbf{\hat{x}}_k^D)^\textup{H}]$, $\mathbf{\Delta}_k=[\textup{diag}(\mathbf{\hat{{\boldsymbol{\delta}}}}_k^1)^\textup{H},...,\textup{diag}(\mathbf{\hat{{\boldsymbol{\delta}}}}_k^D)^\textup{H}]$ and $\mathbf{W}=\mathbf{I}_D\otimes\textup{diag}(\mathbf{w})$. $\textup{diag}(\mathbf{w})$ denotes the diagonal matrix created by the vector $\mathbf{w}$ and $\otimes$ indicates the
Kronecker product. $\mathbf{f}_k=[({\mathbf{f}_k^1})^\textup{H},...,({\mathbf{f}_k^D})^\textup{H}]^\textup{H}$ and $\hat{\mathbf{g}}_k=[({\hat{\mathbf{g}}_k^1})^\textup{H},...,({\hat{\mathbf{g}}_k^D})^\textup{H}]^\textup{H}$ are vectors concatenating the corresponding $D$ vectorized channels. 

\subsubsection{Optimization with ADMM}
The augmented Lagrangian of Eq. (\ref{EQ_RRCF_FT}) is as follows,
\begin{equation}\label{EQ_RRCF_AugLagrangian}
	\begin{split}
		L(\mathbf{f}_k,\hat{\mathbf{g}}_k,\hat{\boldsymbol{\zeta} })=\frac{1}{2}||\mathbf{\hat{y}}-\mathbf{X}_k \hat{\mathbf{g}}_k||_2^2+\frac{\eta }{2}||\mathbf{\Delta} _k^d \hat{\mathbf{g}}_k||_2^2\\
		+\frac{\theta }{2}||\mathbf{W}  \mathbf{f}_k||_2^2+\frac{\tau }{2}|| \mathbf{f}_k-\mathbf{f}_{k-1}||_2^2+\frac{\lambda }{2}||\mathbf{f}_k||_2^2\\
		+\hat{\boldsymbol{\zeta} }^\textup{H}(\hat{\mathbf{g}}_k-(\mathbf{I}_D \otimes \mathbf{FP})\mathbf{f}_k) + \frac{\mu }{2}||\hat{\mathbf{g}}_k-(\mathbf{I}_D \otimes \mathbf{FP})\mathbf{f}_k||_2^2,
	\end{split}
\end{equation}
where $\mu$ is the penalty coefficient and $\hat{\boldsymbol{\zeta} }=[\hat{\boldsymbol{\zeta} }_1^\textup{H},...,\hat{\boldsymbol{\zeta} }_D^\textup{H}]^\textup{H}$ is the Lagrangian vector of size $DN \times 1$ in the frequency domain. Eq. (\ref{EQ_RRCF_AugLagrangian}) is solved iteratively using the ADMM at the $k$th frame. Fortunately, closed form solutions can be found for each of the following subproblems.

\noindent \textbf{Subproblem $\mathbf{f}_k^{\ast}$}

\begin{equation}\label{EQ_RRCF_subproblem_f}
	\begin{split}
		\mathbf{f}_k^{\ast}=\underset{\mathbf{f}_k}{\textup{argmin}}  \left \{     \frac{\theta }{2}||\mathbf{W}  \mathbf{f}_k||_2^2+\frac{\tau }{2}|| \mathbf{f}_k-\mathbf{f}_{k-1}||_2^2+\frac{\lambda }{2}||\mathbf{f}_k||_2^2+ \right. \\ 
		\left. \hat{\boldsymbol{\zeta} }^{\textup{H}}(\hat{\mathbf{g}}_k-(\mathbf{I}_D \otimes \mathbf{FP})\mathbf{f}_k) +  \frac{\mu }{2}||\hat{\mathbf{g}}_k-(\mathbf{I}_D \otimes \mathbf{FP})\mathbf{f}_k||_2^2\right \}\\
		=\left ((\mu + \lambda + \tau)\mathbf{I} + \theta \mathbf{W}^\textup{H}\mathbf{W}  \right )^{-1} \left (\mu \mathbf{g}_k + \boldsymbol{\zeta} + \tau \mathbf{f}_{k-1} \right ),
	\end{split}
\end{equation}
where $\mathbf{g}_k=(\mathbf{I}_D \otimes \mathbf{P}^\textup{T}\mathbf{F}^\textup{H})\hat{\mathbf{g}}_k$ and $\boldsymbol{\zeta}=(\mathbf{I}_D \otimes \mathbf{P}^\textup{T}\mathbf{F}^\textup{H})\hat{\zeta }$, which can be broken into $D$ independent $\mathbf{P}^\textup{T}\mathbf{F}^\textup{H}$ transforms in realization. Since $\mathbf{A}=(\mu + \lambda + \tau)\mathbf{I} + \theta \mathbf{W}^\textup{H}\mathbf{W}$ is a diagonal matrix, its inverse (if it exists) can be computed immediately. In fact, $\mathbf{A}^{-1}$ multiplying by $(\mu \mathbf{g}_k + \boldsymbol{\zeta} + \tau \mathbf{f}_{k-1})$ can be conducted simply by dot division. 

\noindent \textbf{Subproblem $\hat{\mathbf{g}}_k^{\ast}$}

\begin{equation}\label{EQ_RRCF_subproblem_g}
	\begin{split}
		\hat{\mathbf{g}}_k^{\ast}=\underset{\hat{\mathbf{g}}_k}{\textup{argmin}}  \left\{  \frac{1}{2}||\mathbf{\hat{y}}-\mathbf{X}_k \hat{\mathbf{g}}_k||_2^2+\frac{\eta }{2}||\mathbf{\Delta} _k^d \hat{\mathbf{g}}_k||_2^2+ \right. \\
		\left. \hat{\boldsymbol{\zeta} }^\textup{H}(\hat{\mathbf{g}}_k-(\mathbf{I}_D \otimes \mathbf{FP})\mathbf{f}_k) + \frac{\mu }{2}||\hat{\mathbf{g}}_k-(\mathbf{I}_D \otimes \mathbf{FP})\mathbf{f}_k||_2^2  \right \}.
	\end{split}
\end{equation}
It is burden to directly solve Eq. (\ref{EQ_RRCF_subproblem_g}). Fortunately, each entry of $\mathbf{\hat{y}}$, i.e., $\mathbf{\hat{y}}(n)$, depends only on $\mathbf{\hat{x}}_k(n)=[\mathbf{\hat{x}}_k^1(n),...,\mathbf{\hat{x}}_k^D(n)]^\textup{T}$, $\hat{\boldsymbol{\delta}}_k(n)=[\hat{\boldsymbol{\delta}}_k^1(n),...,\hat{\boldsymbol{\delta}}_k^D(n)]^\textup{T}$ and $\hat{\mathbf{g}}_k(n)=[\hat{\mathbf{g}}_k^1(n),...,\hat{\mathbf{g}}_k^D(n)]^\textup{T}$, $n = 1, 2, ..., N$. Therefore, the subproblem $\hat{\mathbf{g}}_k^{\ast}$ can be divided into $N$ smaller problems as follows:
\begin{equation}\label{EQ_RRCF_subsubproblem_g}
	\begin{split}
		\hat{\mathbf{g}}_k^{\ast}(n)=\underset{\hat{\mathbf{g}}_k}{\textup{argmin}}  \left\{  \frac{1}{2}||\mathbf{\hat{y}}(n)-\mathbf{\hat{x}}_k^\textup{H}(n) \hat{\mathbf{g}}_k(n)||_2^2+ \right. \\
		\left. \frac{\eta }{2}||\mathbf{\hat{\boldsymbol{\delta}}} _k^\textup{H}(n) \hat{\mathbf{g}}_k(n)||_2^2+ \hat{\boldsymbol{\zeta} }^\textup{H}(n)(\hat{\mathbf{g}}_k(n)-\hat{\mathbf{f}}_k(n))  \right.\\ 
		\left. +\frac{\mu }{2}||\hat{\mathbf{g}}_k(n)-\hat{\mathbf{f}}_k(n)||_2^2  \right \},
	\end{split}
\end{equation}
where $\hat{\mathbf{f}}_k(n)=[\hat{\mathbf{f}}_k^1(n),...,\hat{\mathbf{f}}_k^D(n)]^\textup{T}$ and $\hat{\mathbf{f}}_k^d=\mathbf{FP}\mathbf{f}_k^d$, i.e., $\hat{\mathbf{f}}_k^d$ is the Fourier transform of padded $\mathbf{f}_k^d$. The solution for $\hat{\mathbf{g}}_k^{\ast}(n)$ is as follows, 
\begin{equation}\label{EQ_RRCF_subsubproblem_g_solution}
	\begin{split}
		\hat{\mathbf{g}}_k^{\ast}(n)=\left (\mu \mathbf{I}_D  + \eta \mathbf{\hat{\boldsymbol{\delta}}} _k(n)\mathbf{\hat{\boldsymbol{\delta}}} _k^\textup{H}(n) + \mathbf{\hat{x}}_k(n)\mathbf{\hat{x}}_k^\textup{H}(n)\right )^{-1}\\
		\left (\mathbf{\hat{y}}(n)\mathbf{\hat{x}}_k(n) + \mu \hat{\mathbf{f}}_k(n) -  \hat{\boldsymbol{\zeta} }(n)\right ).
	\end{split}
\end{equation}
The matrix inversion in Eq. (\ref{EQ_RRCF_subsubproblem_g_solution}) can be got rid of by applying twice the Sherman-Morrison formula \cite{1950Adjustment}, i.e., $(\mathbf{A}+\mathbf{uv}^\textup{H})^{-1}=\mathbf{A}^{-1}-\mathbf{A}^{-1}\mathbf{u(\mathbf{I}+\mathbf{v}^\textup{H}\mathbf{A}^{-1}\mathbf{u})^{-1}}\mathbf{v}^\textup{H}\mathbf{A}^{-1}$, where $\mathbf{u}$ and $\mathbf{v}$ are vectors and $\mathbf{A}$ is an invertible matrix. Specifically, let $\mathbf{A}_1=\mu \mathbf{I}_D  + \eta \mathbf{\hat{\boldsymbol{\delta}}} _k(n)\mathbf{\hat{\boldsymbol{\delta}}} _k^\textup{H}(n)$, $\mathbf{u}_1=\mathbf{v}_1=\mathbf{\hat{x}}_k(n)$ first and then $\mathbf{A}_2=\mu \mathbf{I}_D$, $\mathbf{u}_2=\mathbf{v}_2=\sqrt{\eta } \mathbf{\hat{\boldsymbol{\delta}}} _k(n)$. After further simplification, we have 
\begin{equation}\label{EQ_RRCF_subsubproblem_g_final1}
	\begin{split}
		\hat{\mathbf{g}}_k^{\ast}(n)=\mathbf{A}_1^{-1}(\omega \mathbf{\hat{x}}_k(n) + \mu \hat{\mathbf{f}}_k(n) - \hat{\boldsymbol{\zeta} }(n)),
	\end{split}
\end{equation}
where $\omega $ is defined by
\begin{equation}\label{EQ_RRCF_subsubproblem_g_final2}
	\begin{split}
		\omega =\frac{\mathbf{\hat{y}}(n) + \mu \mathbf{\hat{x}}_k^\textup{H}(n)\mathbf{A}_1^{-1}\hat{\mathbf{f}}_k(n) - \mathbf{\hat{x}}_k^\textup{H}(n)\mathbf{A}_1^{-1}\hat{\boldsymbol{\zeta} }(n)}{(1+\mathbf{\hat{x}}_k(n)^\textup{H}\mathbf{A}_1^{-1}\mathbf{\hat{x}}_k(n))},
	\end{split}
\end{equation}
in which $\mathbf{A}_1^{-1}=\frac{1}{\mu }(\mathbf{I}_D-\frac{\eta \mathbf{\hat{\boldsymbol{\delta}}} _k(n)\mathbf{\hat{\boldsymbol{\delta}}} _k^\textup{H}(n)}{\mu +\eta \mathbf{\hat{\boldsymbol{\delta}}} _k^\textup{H}(n)\mathbf{\hat{\boldsymbol{\delta}}} _k(n)})$. The terms of forms $\mathbf{A}_1^{-1}\mathbf{v}$ and $\mathbf{u}^\textup{H}\mathbf{A}_1^{-1}\mathbf{v}$ in Eq. (\ref{EQ_RRCF_subsubproblem_g_final1}) and Eq. (\ref{EQ_RRCF_subsubproblem_g_final2}), respectively, can be computed effectively by inner product of vectors.

\noindent \textbf{Update of the Lagrangian $\hat{\boldsymbol{\zeta} }$}

The Lagrangian is updated according to:
\begin{equation}\label{EQ_RRCF_subsubproblem_Lagrangians}
	\begin{split}
		\hat{\boldsymbol{\zeta} }_{k}^{(i+1)}=\hat{\boldsymbol{\zeta} }_{k}^{(i)}+\mu(\hat{\mathbf{g}}_k^{\ast(i+1)}-\hat{\mathbf{f}}_k^{\ast(i+1)}))
	\end{split}
\end{equation}
where $\hat{\mathbf{f}}_k^{\ast(i+1)}=(\mathbf{I}_D \otimes \mathbf{FP})\mathbf{f}_k^{\ast(i+1)}$, $\hat{\mathbf{g}}_k^{\ast(i+1)}$ and ${\mathbf{f}}_k^{\ast(i+1)}$ are the current solutions to the two subproblems at iteration $(i+1)$ within ADMM and $\mu$ is set with the scheme $\mu^{i+1}=\textup{min}(\mu_{\textup{max}},\beta\mu ^{(i)})$ \cite{2010Distributed}.
\begin{figure*}[t]
	\centering
	\includegraphics[width=0.75\textwidth]{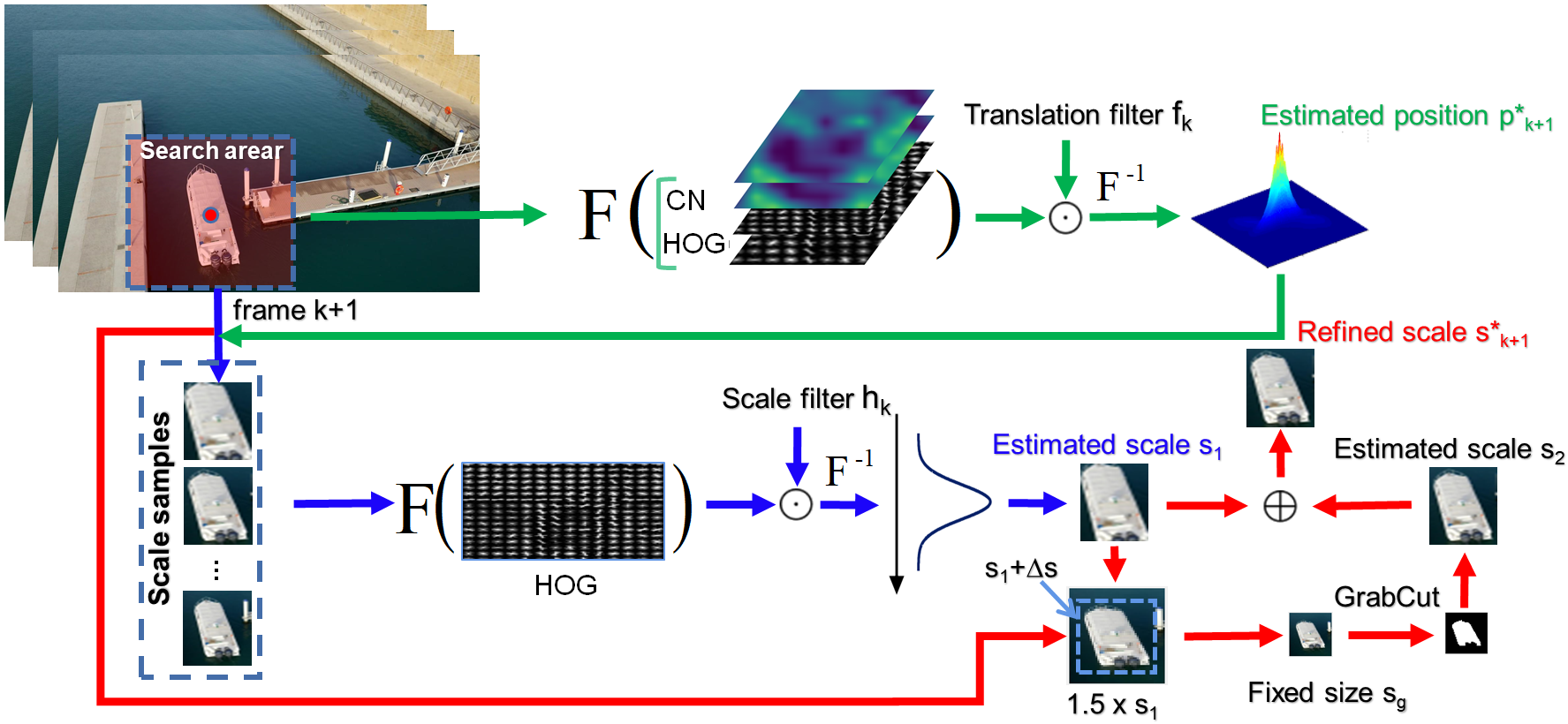}
	\caption{\textbf{Overview of the proposed algorithm.} The tracking task is decomposed into translation and scale estimation. At frame $k+1$, we first infer the target position $\textup{p}^*_{\textup{k}+1}$ using the translation filter $\textup{f}_{\textup{k}}$, and then predict the target scale $\textup{s}_1$ using the scale filter $\textup{h}_{\textup{k}}$. Finally, a segmentation-based scale refinement using the GrabCut is carried out, leading to a refined scale estimate $\textup{s}^*_{\textup{k+1}}$.} \label{fig:tracking_framework}
\end{figure*}
\subsubsection{Update of appearance model}
To improve robustness, the appearance model is adapted online \cite{2015High,2017Learning,huang2019learning} as follows:
\begin{equation}\label{EQ_RRCF_appearance_update}
	\begin{split}
		\mathbf{\hat{x}}_{k}^{model}=(1-\alpha)\mathbf{\hat{x}}_{k-1}^{model}+\alpha\mathbf{\hat{x}}_{k-1}
	\end{split}
\end{equation}
where $\alpha$ is the adaptation rate of the appearance model. $\mathbf{\hat{x}}_{k}^{model}$ instead of $\mathbf{\hat{x}}_{k}$ is then used to solve $\hat{\mathbf{g}}_k^{\ast}$.
\subsubsection{Target localization}
Spatial location of the target in the $(k+1)$th frame is localized by searching for the maximum value of response map $\mathbf{R}_{k+1}$ which is calculated by:
\begin{equation}\label{EQ_RRCF_localization}
	\begin{split}
		\mathbf{R}_{k+1}=\mathbf{F}^{-1}\left (\sum_{d=1}^{D}(\textup{conj}(\mathbf{\hat{z}}_{k+1}^d)\odot \mathbf{\hat{g}}_{k}^d) \right )
	\end{split}
\end{equation}
where $\mathbf{\hat{z}}_{k+1}^d$ denotes the Fourier transform of extracted feature in the $(k+1)$th frame. Operator $\textup{conj}(. )$ denotes the complex conjugate operation.

\subsection{Refine scale estimates with the GrabCut}
As a graph cut \cite{2006Graph} based method, GrabCut is is not only promising to specific images with known information but also effective to the
natural images without any prior knowledge. It uses a Gaussian mixture model \cite{1988Mixture} to estimate the pixel color distribution of the object and background from a user specified bounding box around the segmented object, which is then used to construct a Markov random field \cite{2006Graph} over the pixel labels with an energy function that prefers connected regions having the same label. 

One disadvantage of it is the need for initial user interaction to initialize the segmentation process with a bounding box. Fortunately, in our scenario, we can adapt, by appropriately enlarging, the estimate of SSF to provide such an initial bounding box, which, hopefully, contains background as little as possible with the whole target being inside. Another disadvantage is, GrabCut may produce unacceptable results in the cases of low color contrast between the foreground and the background or high contrast among the foreground itself. Since we intend to exploit the accuracy of the GrabCut for refining, it is very important to refine the scale estimate of SSF only with acceptable segmentation by the GrabCut. Since for an effective SSF-based tracker the IoU (intersection over union) of the scale estimated by SSF with ground truth is supposed to be above a certain value in order to maintain the effectiveness of the appearance model, the IoU between the scale estimated by SSF and that derived from the GrabCut should be above a certain threshold as well if the segmentation was reliable. Therefore, this IoU between the SSF and the GrabCut above a certain threshold is used as the condition in this paper for conducting scale refinement to alleviate accumulating scale errors. The last but not least, although the GrabCut can achieve globally optimal result in polynomial time \cite{2004GrabCut}, it is still slow for UAV tracking if the input to the GrabCut is large. Therefore the input to the GrabCut is resized to a fixed and relatively small size to balance effectiveness and efficiency. 
\begin{figure*}[h]
	\centering
	\subfigure{
		\begin{minipage}[t]{0.24\textwidth}
			\includegraphics[width=1\textwidth]{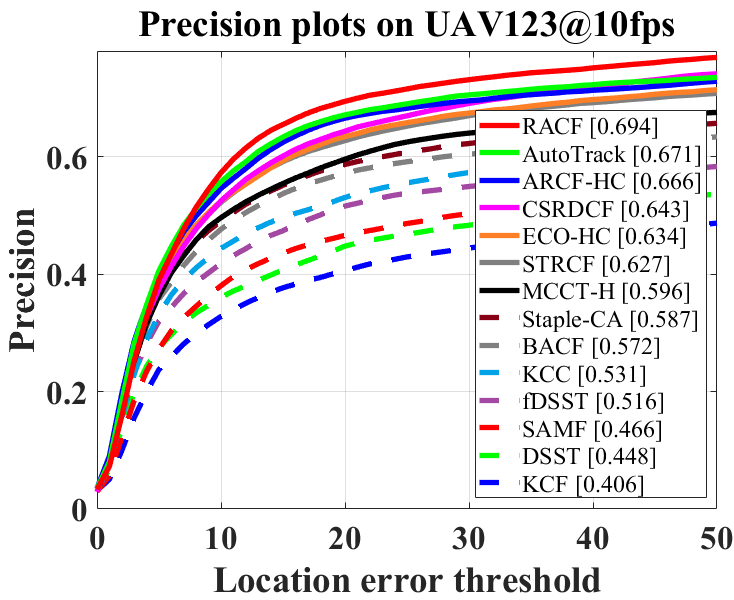}\hspace{0in}
			\includegraphics[width=1\textwidth]{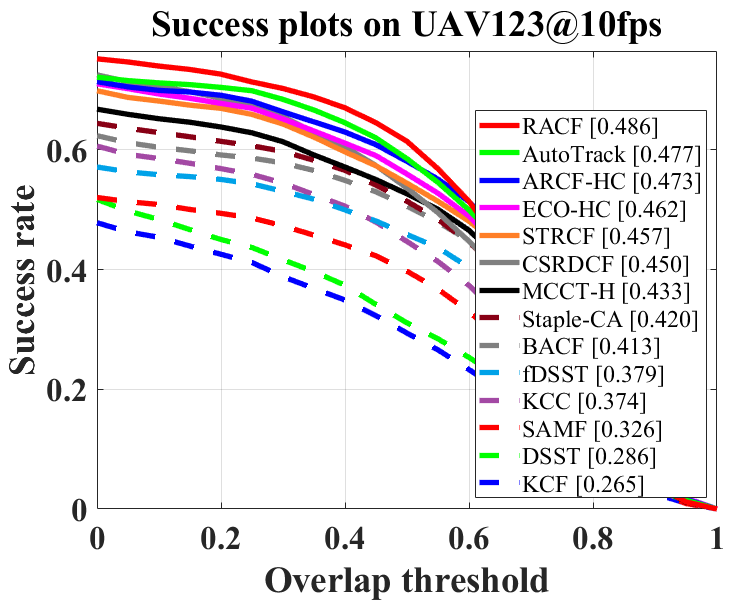}
			\centerline{(a)}
	\end{minipage}}
	\subfigure{
		\begin{minipage}[t]{0.24\textwidth}
			\includegraphics[width=1\textwidth]{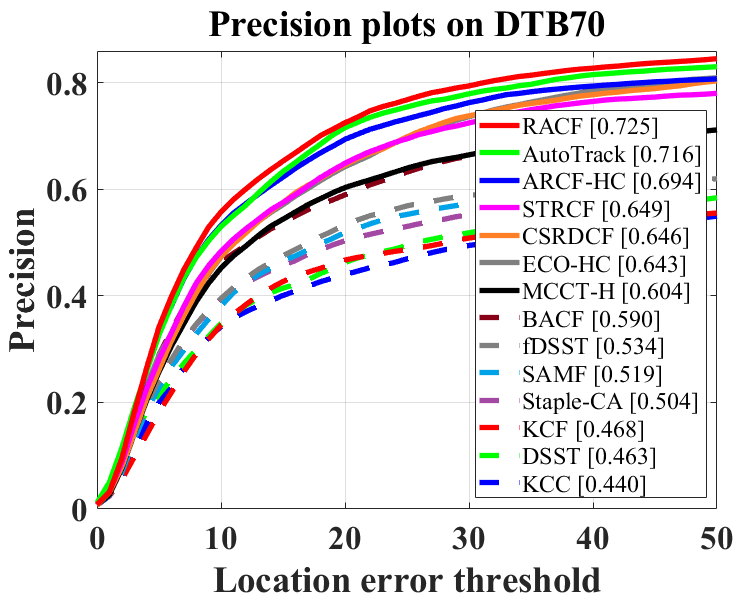}\hspace{0in}
			\includegraphics[width=1\textwidth]{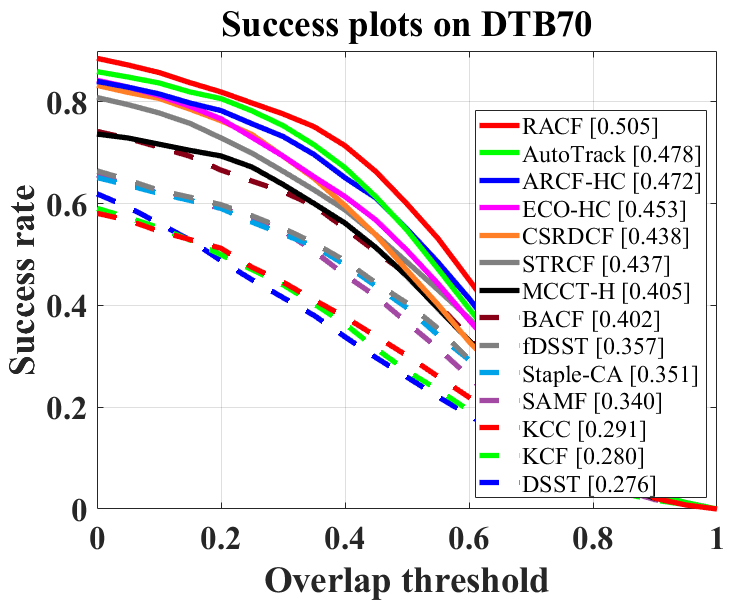}
			\centerline{(b)}
	\end{minipage}}
	\subfigure{
		\begin{minipage}[t]{0.24\textwidth}
			\includegraphics[width=1\textwidth]{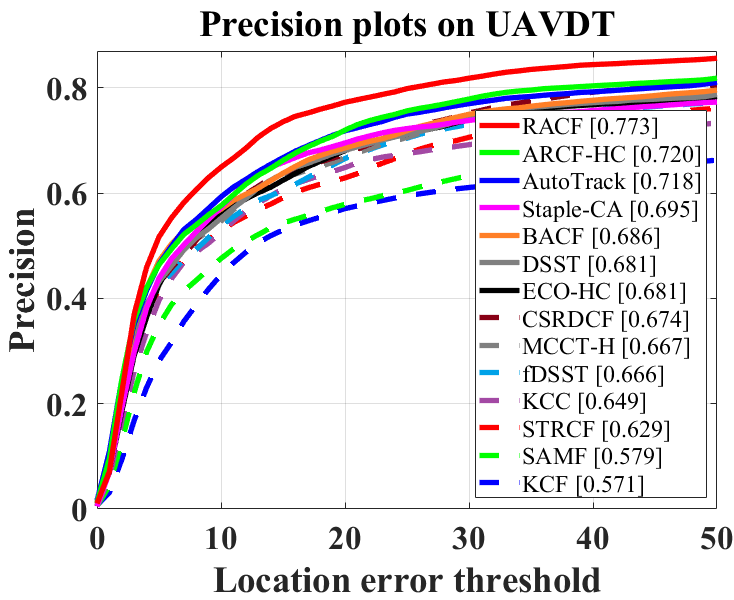}\hspace{0in}
			\includegraphics[width=1\textwidth]{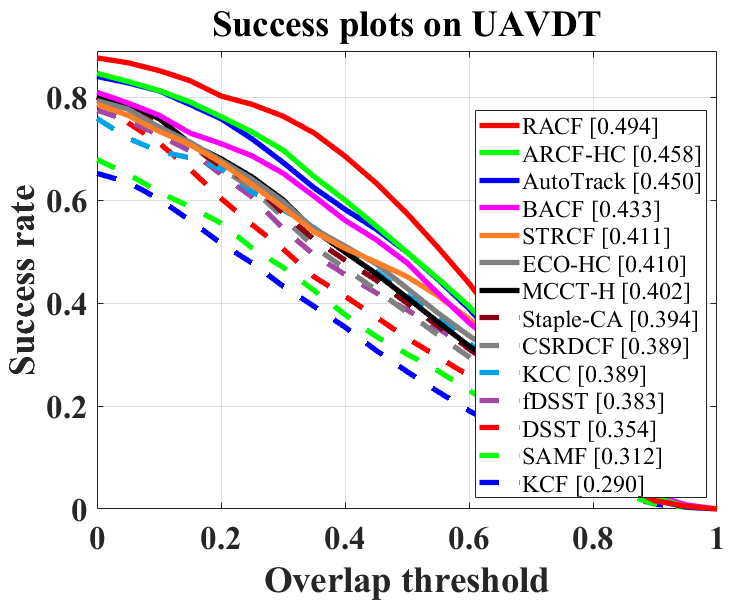}
			\centerline{(c)}
	\end{minipage}}
	\subfigure{
		\begin{minipage}[t]{0.24\textwidth}
			\includegraphics[width=1\textwidth]{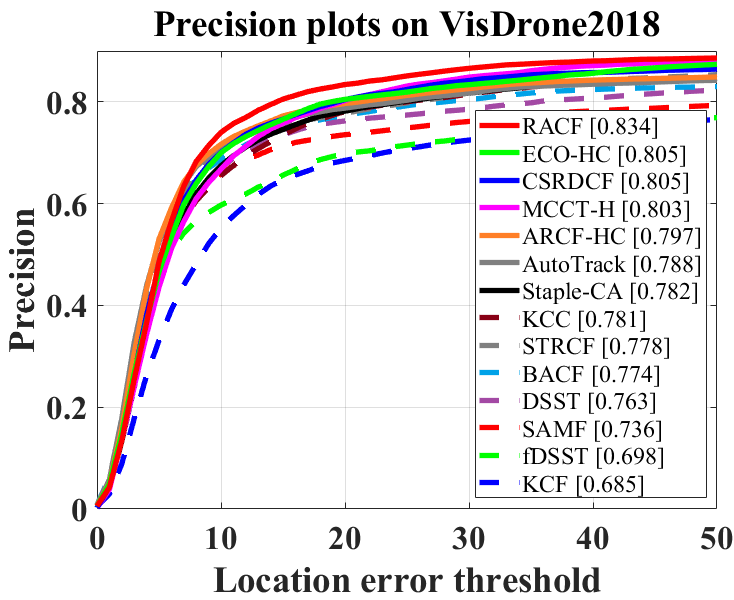}\hspace{0in}
			\includegraphics[width=1\textwidth]{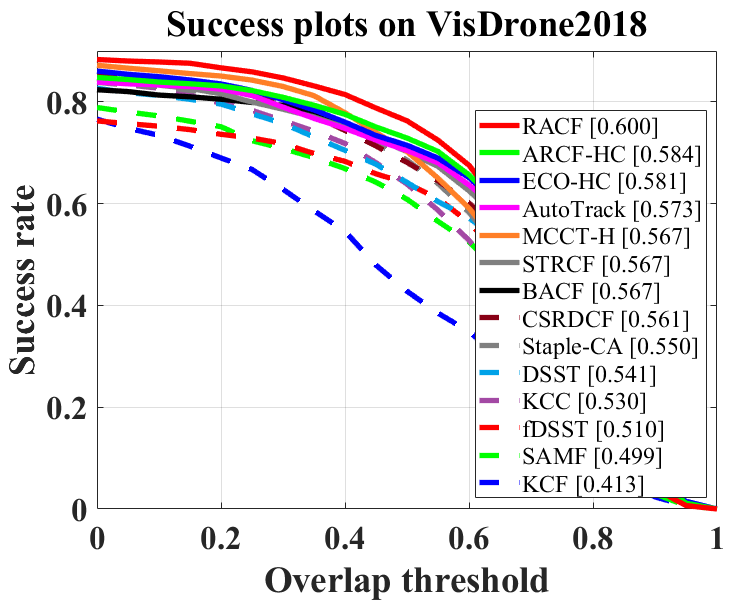}
			\centerline{(d)}
	\end{minipage}}
	\caption{Overall performance of hand-crafted based trackers on (a) UAV123@10fps \cite{2016A} (b) DTB70 \cite{li2017visual}(c) UAVDT \cite{du2018the} and (d) VisDrone2018 (VisDrone2018-test-dev) \cite{wen2018visdrone}. Precision and success rate for one-pass evaluation (OPE) \cite{2013Online} are used for evaluation. The precision at 20 pixels and area under curve (AUC) are used for ranking and marked in the precision plots and success plots respectively.}
	\label{fig_overall_p_s_plots}
\end{figure*}
\subsection{Our tracking framework}
Fig. \ref{fig:tracking_framework} illustrates the overview of the proposed algorithm for visual tracking. The translation filter $\textup{f}_{\textup{k}}$, formulated in the section \ref{subsection_RRCF}, adapts to appearance changes of the target and its surrounding context for estimating translation. The 1D scale filter $\textup{h}_{\textup{k}}$ predicts scale variation of the target is the same as in \cite{danelljan2017discriminative,2017Learning,huang2019learning}. 
The location of the target in the $(k+1)$th frame, denoted by $\textup{p}^*_{\textup{k}+1}$, is estimated by applying $\textup{f}_{\textup{k}}$ that has been updated in the $k$th frame to the detection sample extracted from the current frame at the last position $\textup{p}^*_{\textup{k}}$. Afterwards, $\textup{h}_{\textup{k}}$ is applied at multiple resolutions to estimate the scale of the target size, denoted by $\textup{s}_1$. At last, the proposed scale refinement is carried out. Firstly, an extended example is extracted centered at $\textup{p}^*_{\textup{k+1}}$ with size of $1.5 \textup{s}_1$, which is associated a smaller bounding box of size $\textup{s}_1+\Delta \textup{s}$, $\Delta \textup{s}<0.5\textup{s}_1$, centered also at $\textup{p}^*_{\textup{k+1}}$. The extended example and the bounding box will be scaled to fixed sizes $\textup{s}_g$ and $\left [ \frac{2}{3}(1+\Delta \textup{s}/\textup{s}_1)\textup{s}_g\right ]$, respectively, to make the input image and the initial bounding box for the GrabCut. Then the GrabCut is run to get the binary mask representing the foreground, the minimum bounding box of which, after resized, consists of the estimated scale $\textup{s}_2$ by the GrabCut. Finally, the refined scale is defined by
\begin{equation}\label{EQ_refined_scale}
	\begin{split}
		\textup{s}^*_{\textup{k+1}}=\left\{\begin{matrix}
			\textup{s}_1,& \textup{if} \quad\mathbf{IoU}(\textup{s}_1,\textup{s}_2)>\sigma \\ 
			\textup{s}_2,&\textup{otherwise} 
		\end{matrix}\right. ,
	\end{split}
\end{equation}
where $\mathbf{IoU}(\textup{s}_1,\textup{s}_2)$ computes the IoU between $\textup{s}_1$ and $\textup{s}_2$ and $\sigma$ is a predefined threshold.
\begin{table}[h]
	\centering
	\caption{The proposed trackers with proposed components.}
	\label{tab:Our_Trackers}
	\resizebox{3.2in}{0.28in}{
		\begin{tabular}{cccc}
			\toprule
			& \textbf{RACF}$\ominus \ominus$ & \textbf{RACF}$\ominus$ & \textbf{RACF} \\ \hline\hline
			{Residue-aware regularization}                       & \checkmark       & \checkmark      & \checkmark     \\
			{Spatial \& temporal regularizations} &                 & \checkmark      & \checkmark     \\
			{Scale refinement}                    &                 &                & \checkmark     \\ \hline
		\end{tabular}
	}
	\vspace{-4mm}
\end{table}
\section{Experiments}

In this section, the proposed trackers with different components, summarized in Table \ref{tab:Our_Trackers}, are exhaustively evaluated on four challenging UAV
benchmarks, namely, UAV123@10fps \cite{2016A}, DTB70 \cite{li2017visual}, UAVDT \cite{du2018the} and Vistrone2018 \cite{wen2018visdrone}. UAV123@10fps is designed to investigate the impact of camera
capture speed on tracking performance and constructed by down sampling the UAV123 benchmark \cite{2016A} to 10 FPS from 30FPS. DTB70 composed of 70 UAV sequences primarily addresses the problem of severe UAV motion, but includes as well various cluttered scenes and objects with different sizes. UAVDT is mainly for vehicle tracking with various weather conditions, flying altitudes and camera views. 
Vistrone2018 (VisDrone2018-test-dev) is from the single object
tracking challenge held in conjunction with the european conference on computer vision (ECCV2018), which focuses on evaluating tracking algorithms on drones.
Our feature extractor is the same as in ARCF-HC, i.e., HOG and color name (CN) \cite{huang2019learning}. The evaluation experiments are conducted using MATLAB R2018a on a PC with an i7-7700K processor
(3.7GHz), 32GB RAM and a NVIDIA GTX 960 GPU. For hyper parameters of RACF, we set $\eta=1$, $\theta=0.5$, $\tau=0.01$, $\lambda=0.55$.  The spatial regularization weight is adapted from \cite{li2020asymmetric} with $\rho=1.5$.  The parameters for scale refinement are set with $\textup{s}_g=52$, $\Delta \textup{s}=12$ and $\sigma=0.5$. All parameters settings are kept fixed for all videos in a dataset. ADMM iterations is set to 2. Code is available on: \url{https://github.com/racf2021/racf}



\subsection{Comparison with hand-crafted based trackers}
\begin{figure}[h]
	\centering
	\includegraphics[width=0.475\textwidth,height=0.35\textwidth]{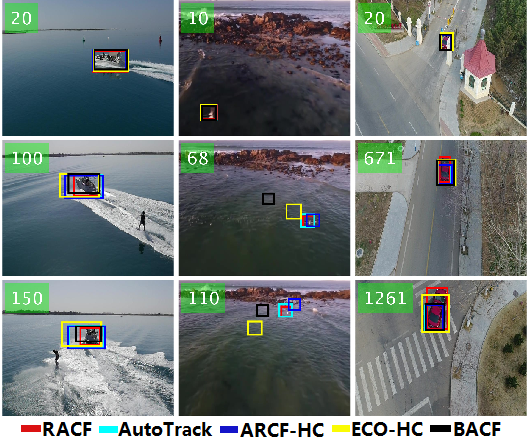}
	\caption{Qualitative evaluation on 3 video sequences from, respectively, UAV123@10fps, DTB70 and UAVDT (i.e. boat7, Gull1 and S1201). The results of AutoTrack, ARCF-HC, ECO-HC, BACF and our RACF are shown with different colors.} \label{fig:visual_examples}
\end{figure}

Thirteen state-of-the-art trackers based on hand-crafted features for comparison are: AutoTrack  \cite{li2020autotrack}, ARCF-HC \cite{huang2019learning}, STRCF \cite{li2018learning}, MCCT-H \cite{wang2018multi}, KCC \cite{2018Kernel}, ECO-HC \cite{danelljan2017eco}, BACF \cite{2017Learning}, Staple-CA \cite{mueller2017context}, CSRDCF  \cite{lukezic2017discriminative}, fDSST \cite{danelljan2016adaptive}, KCF \cite{2015High}, DSST \cite{2014Accurate} and SAMF \cite{li2014a}. The precision and success plots on four benchmarks are shown in Fig. \ref{fig_overall_p_s_plots}. Besides, the average performance in terms of frames per second (FPS)
and precision is displayed in the Table \ref{tab:precision_FPS}. 
\begin{figure*}[h]
	\centering
	\subfigure{
		\begin{minipage}[t]{0.24\textwidth}
			\includegraphics[width=1\textwidth]{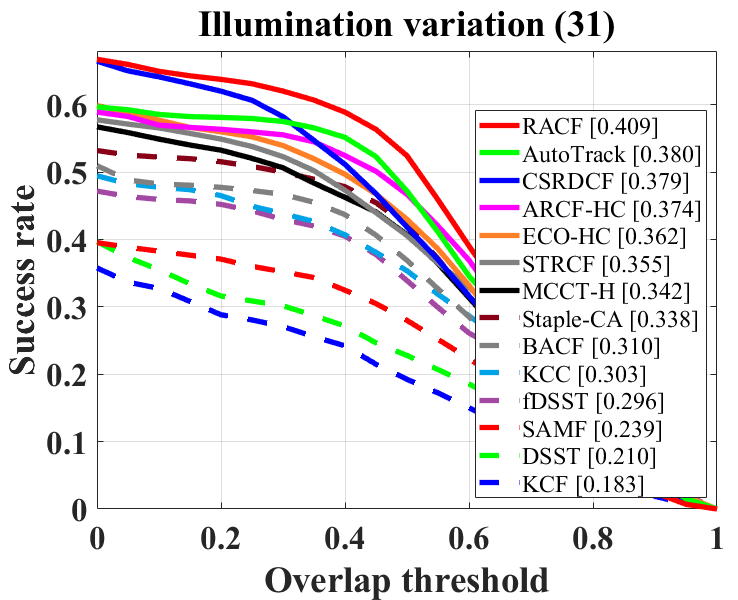}\hspace{0in}
			\includegraphics[width=1\textwidth]{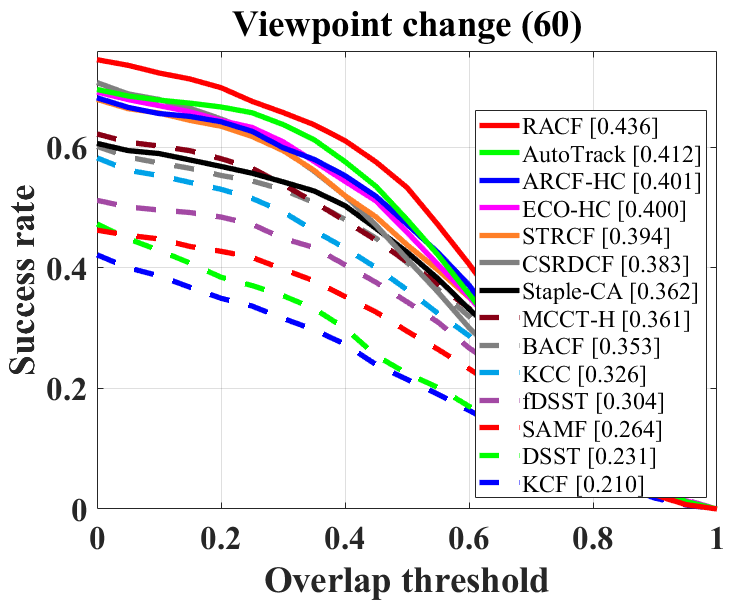}
			\centerline{(a) UAV123@10fps}
	\end{minipage}}
	\subfigure{
		\begin{minipage}[t]{0.24\textwidth}
			\includegraphics[width=1\textwidth]{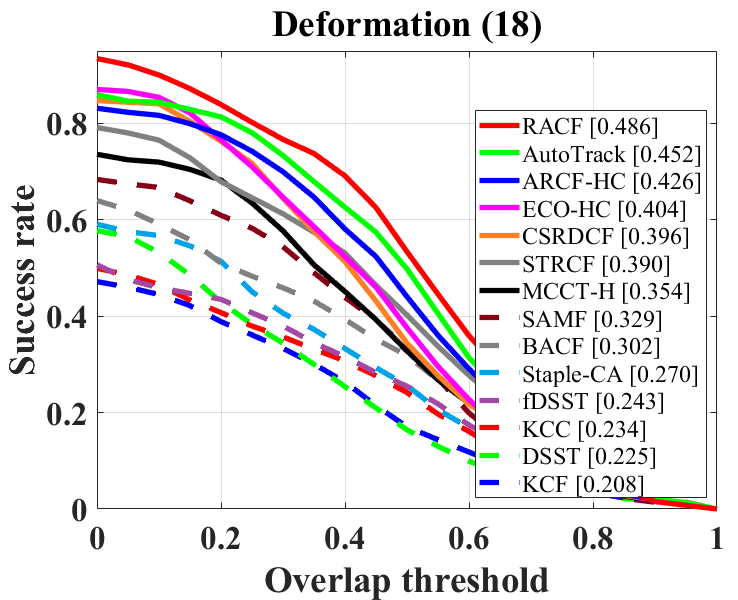}\hspace{0in}
			\includegraphics[width=1\textwidth]{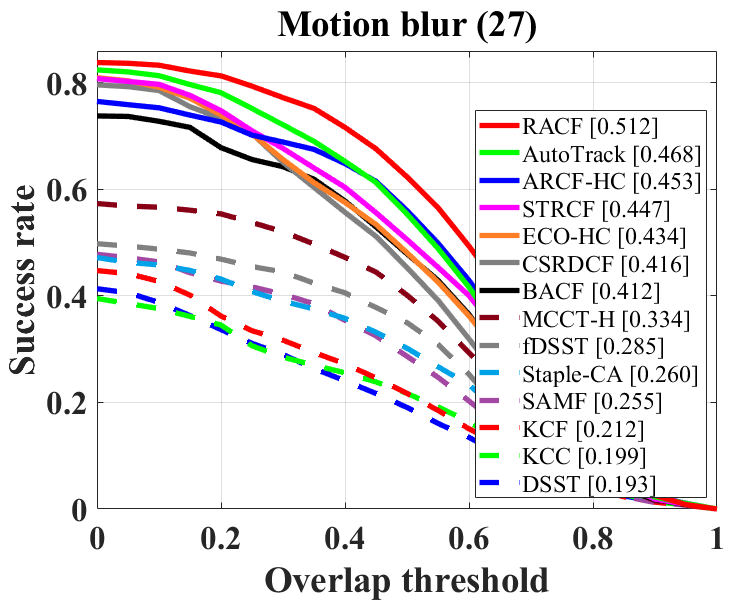}
			\centerline{(b) DTB70}
	\end{minipage}}
	\subfigure{
		\begin{minipage}[t]{0.24\textwidth}
			\includegraphics[width=1\textwidth]{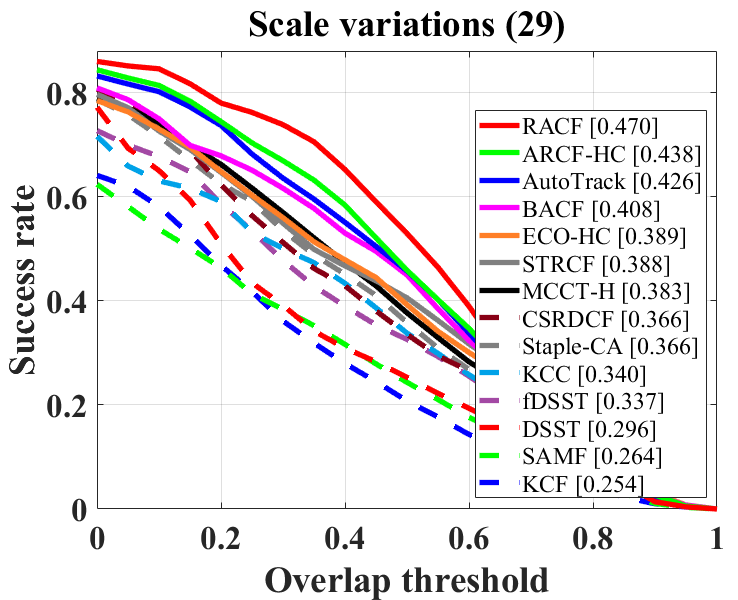}\hspace{0in}
			\includegraphics[width=1\textwidth]{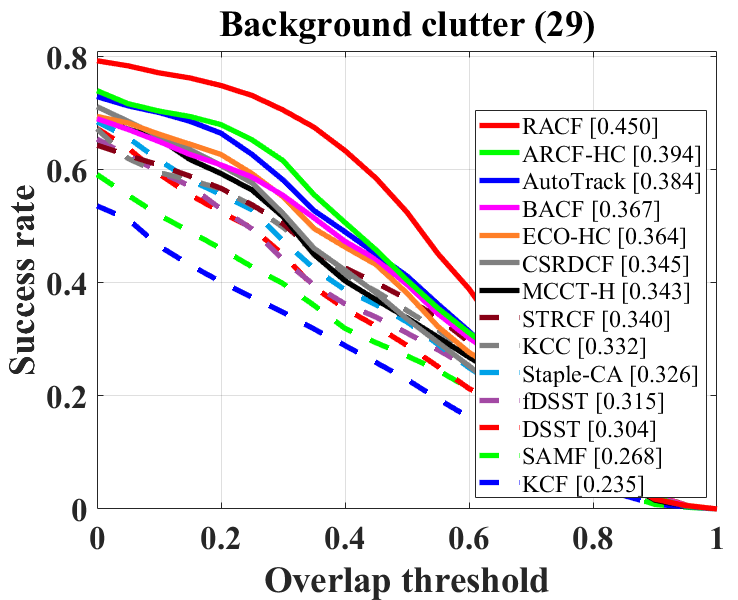}
			\centerline{(c) UAVDT}
	\end{minipage}}
	\subfigure{
		\begin{minipage}[t]{0.24\textwidth}
			\includegraphics[width=1\textwidth]{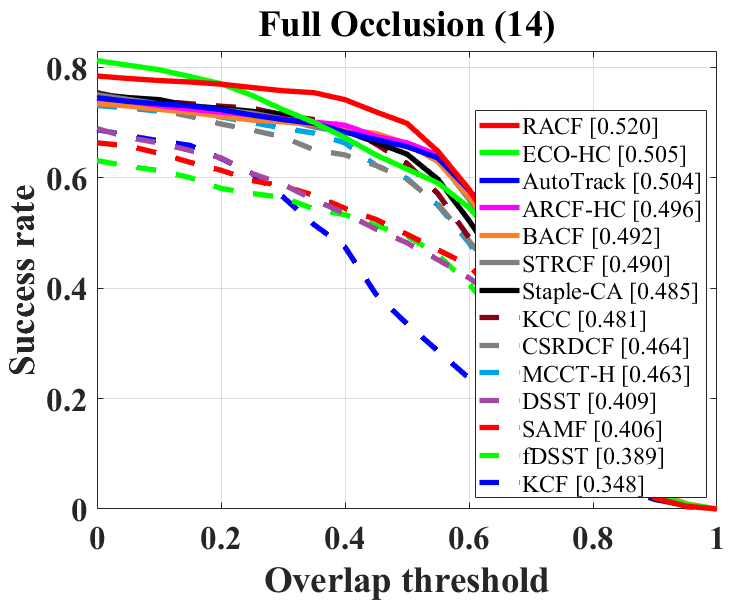}\hspace{0in}
			\includegraphics[width=1\textwidth]{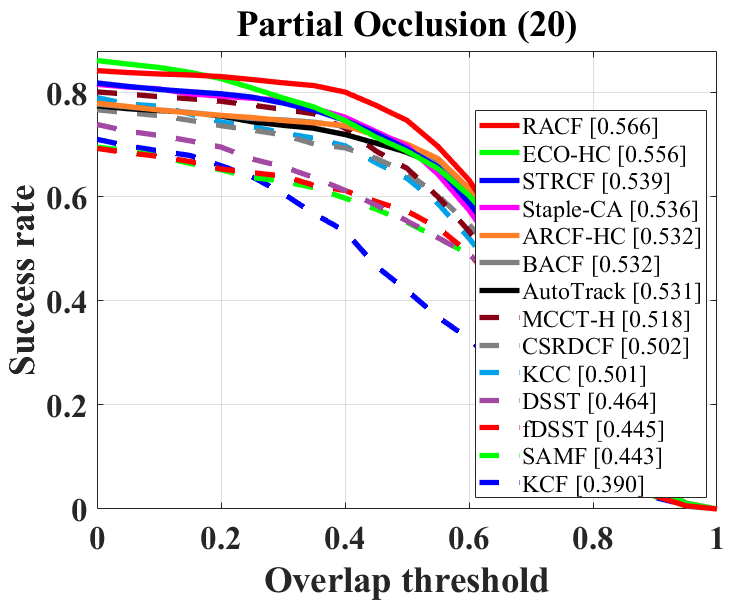}
			\centerline{(d) VisDrone2018}
	\end{minipage}}
	\caption{Attribute-based comparison on illumination variation, viewpoint change, deformation, motion blur, scale variations, background clutter, full occlusion and partial occlusion. }
	\label{fig_attrb_s_plots}
\end{figure*}
\begin{table*}[h]
	\footnotesize
	\centering
	\caption{Average speed (FPS) and precision of hand-crafted based trackers on the four benchmarks. Red, green and blue respectively mean the first, second and third place. All the reported FPSs are evaluated on a single CPU. Noted that RACF$\ominus$ is the best real-time tracker.}
	\label{tab:precision_FPS}
	\resizebox{6.9in}{0.23in}{
		\begin{tabular}{@{}cccccccccccccc    ccc@{}}
			\toprule
			& \textbf{RACF}  &\textbf{RACF}$\ominus$  &\textbf{RACF}$\ominus \ominus$     & AutoTrack\cite{li2020autotrack}                            & ARCF-HC\cite{huang2019learning}                              & KCF\cite{2015High}                                   & DSST\cite{2014Accurate}                                 & BACF\cite{2017Learning} & SAMF\cite{li2014a} & Staple-CA\cite{mueller2017context} & MCCT-H\cite{wang2018multi} & CSRDCF\cite{lukezic2017discriminative} & STRCF\cite{li2018learning} & ECO-HC\cite{danelljan2017eco} & fDSST\cite{danelljan2016adaptive}                                 & KCC\cite{2018Kernel}  \\ \hline\hline
			\textbf{Precision} & {\color[HTML]{FE0000} \textbf{75.7}} &  {\color[HTML]{3531FF} \textbf{74.3}} &{\color[HTML]{009901} \textbf{73.0}}   &72.3 & 71.9 & 53.3                                  & 58.9                                 & 65.3 & 57.5 & 64.2      & 66.8   & 69.2   & 67.1  & 68.8   & 60.4                                  & 60.0 \\
			\textbf{FPS}      & 26.3    & 36.2        &36.3                     & 43.2                                 & 25.2                                 & {\color[HTML]{FE0000} \textbf{545.8}} & {\color[HTML]{009901} \textbf{88.3}} & 45.1 & 9.6  & 52.7      & 50.9   & 10.6   & 23.0  & 62.1   & {\color[HTML]{3531FF} \textbf{164.7}} & 35.2 \\ \hline
		\end{tabular}
	}
\end{table*}
\\\indent\textbf{Overall performance evaluation:} Fig. \ref{fig_overall_p_s_plots} shows the overall performance of RACF with the competing trackers on the four benchmarks. As can be seen, RACF outperforms all other trackers on all benchmarks.
Specifically, on UAV123@10fps and DTB70, RACF outperforms the second tracker AutoTrack in (precision, AUC) with gains of (2.3\%, 0.9\%) and (0.9\%, 2.7\%) respectively. On UAVDT, RACF surpasses the second place, i.e, ARCF-HC, by a significant gain of (5.3\%,3.6\%).
RACF also achieved the best performance on Vistrone2018, followed by ECO-HC with a gap of (2.9\%, 1.6\%). In terms of speed, we evaluate the average FPS of RACF, RACF$\ominus$ and RACF$\ominus\ominus$ with the competing trackers on the four benchmarks, the FPSs along with the average precisions are shown in Table \ref{tab:precision_FPS}. As can be seen, RACF$\ominus\ominus$ has already outperformed state-of-the-art trackers in precision and RACF$\ominus$ is the best real-time tracker (with a speed of $>$30FPS) on CPU. It also shows that RACF is near real-time even with scale refinement, and it makes little difference whether the spatial and temporal regularizations are used or not because $\tau\mathbf{I}$ and $\theta \mathbf{W}^\textup{H}\mathbf{W}$ in (\ref{EQ_RRCF_subproblem_f}) are both diagonal matrices. 
\\\indent\textbf{Attribute-based evaluation:}
The proposed RACF outperforms other hand-crafted based trackers in most attributes defined respectively in the four benchmarks. Examples of success plots are shown in Fig. \ref{fig_attrb_s_plots}. In the situations of deformation, viewpoint change and scale variations, RACF demonstrates a significant improvement over other trackers because the effectiveness of the proposed scale refinement in the corresponding benchmarks. {All results are displayed in the supplementary materials}.
\\\indent\textbf{Qualitative evaluation:}
Some qualitative tracking results of RACF and four top trackers are shown in Fig. \ref{fig:visual_examples}. It can be seen that the scale estimates of RACF are more accurate in these examples with challenging viewpoint change, deformation and scale variations, justifying the effectiveness of the proposed method for refining scale estimates.
\subsection{Comparison with deep-based trackers}
The proposed RACF is also compared with fifteen state-of-the-art
deep trackers, i.e., KYS \cite{bhat2020know}, D3S \cite{lukezic2020d3s}, SiamR-CNN \cite{voigtlaender2020siam}, PrDiMP18 \cite{danelljan2020probabilistic}, ASRCF \cite{dai2019visual}, UDT+ \cite{wang2019unsupervised},
TADT \cite{li2019target}, DeepSTRCF \cite{li2018learning},  MCCT \cite{2018Multi}, DSiam \cite{2017Learning}, ECO \cite{danelljan2017eco}, ADNet \cite{2017Action}, CFNet \cite{2017End}, MCPF \cite{2017Multi} and CREST \cite{song2017crest}. RACF achieves the second best precision and its CPU speed surpasses most GPU speeds on the UAVDT benchmark as shown in Table \ref{tab:comparision_with_deep_trackers}.

\begin{table}[]
	\small
	\centering
	\caption{Precision and speed (FPS) comparison between RACF and
		deep-based trackers on UAVDT \cite{du2018the}. * means GPU speed. Red, green and blue respectively mean the first, second and third place.}
	\label{tab:comparision_with_deep_trackers}
	\resizebox{3.2in}{0.6in}{
		\begin{tabular}{@{}ccc|ccc@{}}
			\toprule
			Tracker                                     & Precision                            & FPS                                  & Tracker                      & Precision                            & FPS                                  \\ \midrule
			\hline
			\textbf{RACF}                               & {\color[HTML]{3531FF} \textbf{77.3}} & {\color[HTML]{009901} \textbf{28.0}} & KYS \cite{bhat2020know}     & {\color[HTML]{FE0000} \textbf{79.8}} & 18.6*                                 \\
			PrDiMP18 \cite{danelljan2020probabilistic} & {\color[HTML]{009901} \textbf{75.1}} & {\color[HTML]{3531FF} \textbf{30.5*}} & D3S \cite{lukezic2020d3s}   & 72.2                                 & {\color[HTML]{009901} \textbf{28.0*}} \\
			SiamR-CNN \cite{voigtlaender2020siam}      & 66.5                                 & 4.6*                                  & TADT \cite{li2019target}    & 67.7                                 & {\color[HTML]{000000} 19.1*}          \\
			ASRCF \cite{dai2019visual}                 & 70.0                                 & 14.2*                                 & MCCT \cite{2018Multi}       & 67.1                                 & 5.1*                                  \\
			UDT+ \cite{wang2019unsupervised}           & 69.7                                 & {\color[HTML]{FE0000} \textbf{35.5*}}                                 & ADNet \cite{2017Action}     & 68.3                                 & 4.7*                                  \\
			DeepSTRCF \cite{li2018learning}            & 66.7                                 & 3.9*                                  & ECO \cite{danelljan2017eco} & 72.1                                 & 27.9* \\
			MCPF \cite{2017Multi}                      & 66.0                                 & 0.4*                                  & CFNet \cite{2017End}        & 68.1                                 & 24.2*                                 \\
			DSiam \cite{2017Learning}                  & 70.4                                 & 9.3*                                  & CREST \cite{song2017crest}  & 64.9                                 & 2.5*                                 \\ 
			\hline
		\end{tabular}
	}
\end{table}

\subsection{Ablation study}
\begin{table*}[]
	\centering
	\small
	\caption{Illustration of how success rate on DTB70 varies  with the number of ADMM iterations in RACF$\ominus \ominus$ and the two baselines, i.e., BACF \cite{2017Learning} and ARCF-HC \cite{huang2019learning}. The best success rates and the ones corresponding to default parameter settings are shown in red and underline, respectively.}
	\label{tab:Iters_about_Optm}
	\resizebox{6.5in}{0.35in}{
		\begin{tabular}{@{}cccccccccccccc@{}}
			\toprule
			\diagbox[trim=l]{\textbf{Tracker}}{\textbf{Iter.}}   & 1      & 2                                            & 3                                      & 4      & 5                     & 6      & 7      & 8      & 9      & 10     & 15                                     & 20     & 25     \\ 
			\hline\hline
			BACF\cite{2017Learning} & 23.113 & {\ul \textbf{40.201}}                        & {\color[HTML]{FE0000} \textbf{41.016}} & 40.864 & 40.695                & 40.866 & 40.845 & 40.577 & 40.555 & 40.72  & 40.773                                 & 40.433 & 40.615 \\
			ARCF-HC\cite{huang2019learning} & 29.609 & 46.262                                       & 45.785                                 & 46.806 & {\ul \textbf{47.171}} & 47.103 & 47.136 & 47.255 & 47.261 & 47.254 & {\color[HTML]{FE0000} \textbf{47.482}} & 46.357 & 46.772 \\
			\textbf{RACF}$\ominus \ominus$ & 27.550  & {\color[HTML]{FE0000} {\ul \textbf{48.246}}} & 48.184                                 & 47.893 & 47.893                & 47.893 & 47.898 & 47.898 & 47.898 & 47.898 & 47.903                                 & 47.872 & 47.871 \\ \hline
		\end{tabular}
	}
\end{table*}
\begin{table*}[h]
	\centering
	\small
	\caption{AUCs and precisions on the four benchmarks. PRC is short for precision, and SR represents the scale refinement component. Red, green and blue respectively mean the first, second and third place.}
	\label{tab:Ablation_on_SR_ST}
	\resizebox{6.5in}{0.5in}{
		\begin{tabular}{@{}ccccccccccc@{}}
			\toprule
			& \multicolumn{2}{c}{UAV123@10fps}                                                 & \multicolumn{2}{c}{DTB70}                                                             & \multicolumn{2}{c}{UAVDT}                                                             & \multicolumn{2}{c}{Vistrone2018}   
			& \multicolumn{2}{c}{Avg.}                                                     \\  \cmidrule(r){2-3}   \cmidrule(r){4-5}  \cmidrule(r){6-7}  \cmidrule(r){8-9} \cmidrule(r){10-11} 
			\multirow{-2}{*}{\textbf{Methods}} & AUC                                       & RPC                                  & AUC                                       & RPC                                       & AUC                                       & RPC                                       &
			AUC                                       & RPC                                       & AUC                                       & RPC                                       \\ \hline \hline
			ECO-HC\cite{danelljan2017eco} w/wo SR     & 46.2/47.7                                 & 63.4/66.0                            & 45.3/46.0                                 & 64.3/64.6                                 & 41.0/46.5                                 & 68.1/72.7                                 & 58.1/58.8                                 & 80.5/81.2                                 &
			47.7/49.8                               &
			69.1/71.1\\
			ARCF-HC\cite{huang2019learning} w/wo SR    & 47.3/47.5                                 & 66.6/66.7                            & 47.2/{\color[HTML]{32CB00} \textbf{48.9}} & 69.4/69.5                                 & 45.8/{\color[HTML]{3166FF} \textbf{49.3}} & 72.0/{\color[HTML]{3166FF} \textbf{77.7}} & 58.4/{\color[HTML]{3166FF} \textbf{59.7}} & 79.7/{\color[HTML]{32CB00} \textbf{81.3}} &
			49.7/{\color[HTML]{3166FF} \textbf{51.4 }}                              &
			71.9/73.8\\
			AutoTrack\cite{li2020autotrack}w/wo SR  & 47.7/{\color[HTML]{3166FF} \textbf{48.5}} & 67.1/67.8                            & 47.8/{\color[HTML]{3166FF} \textbf{49.3}} & 71.6/{\color[HTML]{3166FF} \textbf{71.7}} & 45.0/{\color[HTML]{32CB00} \textbf{48.2}} & 71.8/75.2                                 & 57.3/58.8                                 & 78.8/80.8                                 &
			49.5/{\color[HTML]{32CB00} \textbf{51.2 }}                              &
			72.3/{\color[HTML]{32CB00} \textbf{73.9 }}\\
			\cdashline{1-11} 
			\textbf{RACF} $\ominus \ominus$             & {\color[HTML]{32CB00} \textbf{47.8}}      & {\color[HTML]{32CB00} \textbf{68.6}} & 48.2                                      & 70.3                                      & 45.6                                      & 73.8                                      & 57.3                                      & 79.3                                      &
			49.7                                     &
			73.0\\
			\textbf{RACF} $\ominus \quad$              & {\color[HTML]{32CB00} \textbf{47.8}}      & {\color[HTML]{3166FF} \textbf{69.1}} & 48.5                                      & {\color[HTML]{32CB00} \textbf{70.7}}      & 46.6                                      & {\color[HTML]{32CB00} \textbf{75.3}}      & {\color[HTML]{32CB00} \textbf{59.2}}      & {\color[HTML]{3166FF} \textbf{82.1}}      &
			50.5                                                           &
			{\color[HTML]{3166FF} \textbf{74.3}}\\
			\textbf{RACF}$\qquad$               & {\color[HTML]{FE0000} \textbf{48.6}}      & {\color[HTML]{FE0000} \textbf{69.4}} & {\color[HTML]{FE0000} \textbf{50.5}}      & {\color[HTML]{FE0000} \textbf{72.5}}      & {\color[HTML]{FE0000} \textbf{49.4}}      & {\color[HTML]{FE0000} \textbf{77.3}}      & {\color[HTML]{FE0000} \textbf{60.0}}      & {\color[HTML]{FE0000} \textbf{83.4}}      &
			{\color[HTML]{FE0000} \textbf{52.1}}                                     &
			{\color[HTML]{FE0000} \textbf{75.6}}\\ \hline
		\end{tabular}
	}
\end{table*}
\textbf{Residue-aware regularization:} To see how the number of ADMM iterations impacts the success rate of BACF, ARCF-HC and RACF$\ominus \ominus$, respectively, we evaluate the three trackers on DTB70 with varied ADMM iterations. The variations of success rate with respect to the number of ADMM iterations are shown in Table \ref{tab:Iters_about_Optm}. As can be seen, RACF$\ominus \ominus$ obtains the highest success rate at iteration 2, BACF at iteration 3 and ARCF-HC at iteration 15, justifying that RACF$\ominus \ominus$ converges faster in average. We can also see that the fluctuations after iteration 2 are basically smaller in RACF$\ominus \ominus$ than in the others. To study the impact of $\eta$, the penalty of residue-aware regularization, we show in Tabel \ref{tab:Iter_of_Base} the success rates of RACF$\ominus \ominus$ on DTB70 with respect to both $\eta$ and the number of ADMM iterations. As shown, RACF$\ominus \ominus$ achieves the highest success rate when $\eta=1.0$ at iteration 2. Notice that for each value of $\eta$ the highest success rate is obtained at iteration either 2 or 3, which conforms the fast convergence of the method.
\begin{table}[]
	\caption{Illustration of how success rate on DTB70 varies  with $\eta$ and the number of ADMM iterations in RACF$\ominus \ominus$. The best success rates are shown in red.}
	\label{tab:Iter_of_Base}
	\resizebox{3.2in}{0.75in}{
		\begin{tabular}{@{}ccccccccc@{}}
			\toprule
			\textbf{}\diagbox[trim=l]{\textbf{Iter.}}{$\mathbf{\eta}$} & 0.2                                    & 0.4                                    & 0.6                                    & 0.8                                    & 1.0                                    & 1.2                                    & 1.4                                    & 1.6                                    \\ \hline\hline
			1         & 28.782                                 & 27.933                                 & 27.927                                 & 27.083                                 & 27.550                                 & 27.342                                 & 27.672                                 & 28.295                                 \\
			2         & 46.590                                 & {\color[HTML]{FF0000} \textbf{46.977}} & {\color[HTML]{FF0000} \textbf{47.903}} & {\color[HTML]{FF0000} \textbf{48.098}} & {\color[HTML]{FF0000} \textbf{48.246}} & 46.971                                 & {\color[HTML]{FF0000} \textbf{47.031}} & {\color[HTML]{FF0000} \textbf{47.203}} \\
			3         & {\color[HTML]{FF0000} \textbf{47.039}} & 46.879                                 & 47.712                                 & 47.544                                 & 48.184                                 & {\color[HTML]{FF0000} \textbf{47.603}} & 46.595                                 & 46.572                                 \\
			4         & 47.018                                 & 46.843                                 & 47.752                                 & 47.522                                 & 47.893                                 & 47.190                                 & 46.651                                 & 46.595                                 \\
			5         & 47.018                                 & 46.843                                 & 47.752                                 & 47.522                                 & 47.893                                 & 47.190                                 & 46.651                                 & 46.595                                 \\
			6         & 47.018                                 & 46.806                                 & 47.752                                 & 47.522                                 & 47.893                                 & 47.190                                 & 46.700                                 & 46.595                                 \\
			7         & 47.018                                 & 46.806                                 & 47.752                                 & 47.521                                 & 47.898                                 & 47.190                                 & 46.700                                 & 46.595                                 \\
			8         & 47.018                                 & 46.806                                 & 47.752                                 & 47.521                                 & 47.898                                 & 47.190                                 & 46.700                                 & 46.595                                 \\
			9         & 47.018                                 & 46.806                                 & 47.752                                 & 47.521                                 & 47.898                                 & 47.189                                 & 46.700                                 & 46.587                                 \\
			10        & 47.018                                 & 46.796                                 & 47.839                                 & 47.521                                 & 47.898                                 & 47.189                                 & 46.700                                 & 46.587                                 \\
			15        & 47.035                                 & 46.773                                 & 47.843                                 & 47.521                                 & 47.903                                 & 47.152                                 & 46.660                                 & 46.591                                 \\
			20        & 47.029                                 & 46.773                                 & 47.640                                 & 47.526                                 & 47.872                                 & 47.155                                 & 46.660                                 & 46.544                                 \\
			25        & 47.029                                 & 46.747                                 & 47.590                                 & 47.540                                 & 47.871                                 & 47.155                                 & 46.680                                 & 46.541                                 \\ \hline
		\end{tabular}
	}
\end{table}

\indent\textbf{Scale refinement, spatial and temporal regularizations:} We incorporate the scale refinement (SR) component into three sate-of-the-art trackers, i.e, ECO-HC, ARCF-HC and AutoTrack, to validate its effectiveness and generality. The AUC and PRC, short for precision, on the four benchmarks are shown in Table \ref{tab:Ablation_on_SR_ST}. Equipped with SR, ECO-HC, ARCF-HC and AutoTrack acquire in average, respectively, gains of (2.1\%, 2.0\%), (1.7\%, 1.9\%) and (1.7\%, 1.6\%) in (AUC, PRC). The most significant improvements, specifically (5.5\%, 4.6\%), (3.5\%, 5.7\%) and (3.2\%, 3.4\%), are observed on UAVDT, for whose foreground and background are basically cars and roads and the GrabCut works well in these cases. The AUC and PRC of the proposed trackers are also shown in Table \ref{tab:Ablation_on_SR_ST} beneath the dash line. As can be seen, in average, the SR component makes a gain of (1.6\%, 1.3\%) in (AUC, PRC), while the spatial and temporal regularizations make a gain of (0.8\%, 1.3\%), justifying the effectiveness of both components.

\section{Conclusion}

In this work, we proposed and evaluated the residue-aware correlation filters and the method of refining scale estimates with the GrabCut. The proposed RACF surpasses all sate-of-the-art hand-crafted based trackers in terms of precision and success rate in all the four UAV benchmarks and is comparable as well to many state-of-the-art deep-based trackers on UAVDT. The proposed scale refinement component can be easily incorporated into any tracking method with discriminative scale estimation to improve precision and accuracy.
{\small
	\bibliographystyle{ieee_fullname}
	\bibliography{egbib}

\begin{thebibliography}{10}\itemsep=-1pt

\bibitem{2019Patch}
Xiaofei~Du A, Maximilian~Allan B, Sebastian~Bodenstedt C, Maier~Hein D,
  Stefanie~Speidel C, Alessio~Dore E, and Danail~Stoyanov A.
\newblock Patch-based adaptive weighting with segmentation and scale (pawss)
  for visual tracking in surgical video.
\newblock {\em Medical Image Analysis}, 57:120--135, 2019.

\bibitem{aeschliman2010a}
Chad {Aeschliman}, Johnny {Park}, and Avinash~C. {Kak}.
\newblock A probabilistic framework for joint segmentation and tracking.
\newblock In {\em 2010 IEEE Computer Society Conference on Computer Vision and
  Pattern Recognition}, pages 1371--1378, 2010.

\bibitem{balduzzi2017the}
David {Balduzzi}, Marcus {Frean}, Lennox {Leary}, J~P {Lewis}, Kurt Wan-Duo
  {Ma}, and Brian {McWilliams}.
\newblock The shattered gradients problem: if resnets are the answer, then what
  is the question?
\newblock In {\em ICML'17 Proceedings of the 34th International Conference on
  Machine Learning - Volume 70}, pages 342--350, 2017.

\bibitem{bhat2020know}
Goutam {Bhat}, Martin {Danelljan}, Luc~Van {Gool}, and Radu {Timofte}.
\newblock Know your surroundings: Exploiting scene information for object
  tracking.
\newblock In {\em ECCV (23)}, pages 205--221, 2020.

\bibitem{2016Target}
Adel Bibi, Matthias Mueller, and Bernard Ghanem.
\newblock Target response adaptation for correlation filter tracking.
\newblock 2016.

\bibitem{bolme2010visual}
David~S Bolme, J~Ross Beveridge, Bruce~A Draper, and Yui~Man Lui.
\newblock Visual object tracking using adaptive correlation filters.
\newblock In {\em 2010 IEEE computer society conference on computer vision and
  pattern recognition}, pages 2544--2550. IEEE, 2010.

\bibitem{2010Distributed}
Stephen Boyd, Neal Parikh, Eric Chu, Borja Peleato, and Jonathan Eckstein.
\newblock Distributed optimization and statistical learning via the alternating
  direction method of multipliers.
\newblock {\em Foundations \& Trends in Machine Learning}, 3(1):1--122, 2010.

\bibitem{2006Graph}
Yuri Boykov and Gareth Funka-Lea.
\newblock Graph cuts and efficient n-d image segmentation.
\newblock {\em International Journal of Computer Vision}, 70(2):109--131, 2006.

\bibitem{briggs2000multigrid}
William~L Briggs, Van~Emden Henson, and Steve~F McCormick.
\newblock {\em A multigrid tutorial}.
\newblock SIAM, 2000.

\bibitem{chatfield2011the}
Ken {Chatfield}, Victor~S. {Lempitsky}, Andrea {Vedaldi}, and Andrew
  {Zisserman}.
\newblock The devil is in the details: an evaluation of recent feature encoding
  methods.
\newblock In {\em British Machine Vision Conference 2011}, pages 1--12, 2011.

\bibitem{chen2020state}
Xi {Chen}, Zuoxin {Li}, Ye {Yuan}, Gang {Yu}, Jianxin {Shen}, and Donglian
  {Qi}.
\newblock State-aware tracker for real-time video object segmentation.
\newblock In {\em 2020 IEEE/CVF Conference on Computer Vision and Pattern
  Recognition (CVPR)}, pages 9384--9393, 2020.

\bibitem{dai2019visual}
Kenan {Dai}, Dong {Wang}, Huchuan {Lu}, Chong {Sun}, and Jianhua {Li}.
\newblock Visual tracking via adaptive spatially-regularized correlation
  filters.
\newblock In {\em 2019 IEEE/CVF Conference on Computer Vision and Pattern
  Recognition (CVPR)}, pages 4670--4679, 2019.

\bibitem{danelljan2017eco}
Martin {Danelljan}, Goutam {Bhat}, Fahad~Shahbaz {Khan}, and Michael
  {Felsberg}.
\newblock Eco: Efficient convolution operators for tracking.
\newblock In {\em 2017 IEEE Conference on Computer Vision and Pattern
  Recognition (CVPR)}, pages 6931--6939, 2017.

\bibitem{danelljan2020probabilistic}
Martin {Danelljan}, Luc~Van {Gool}, and Radu {Timofte}.
\newblock Probabilistic regression for visual tracking.
\newblock In {\em 2020 IEEE/CVF Conference on Computer Vision and Pattern
  Recognition (CVPR)}, pages 7183--7192, 2020.

\bibitem{danelljan2015learning}
Martin {Danelljan}, Gustav {Hager}, Fahad~Shahbaz {Khan}, and Michael
  {Felsberg}.
\newblock Learning spatially regularized correlation filters for visual
  tracking.
\newblock In {\em 2015 IEEE International Conference on Computer Vision
  (ICCV)}, pages 4310--4318, 2015.

\bibitem{danelljan2016adaptive}
Martin {Danelljan}, Gustav {Hager}, Fahad~Shahbaz {Khan}, and Michael
  {Felsberg}.
\newblock Adaptive decontamination of the training set: A unified formulation
  for discriminative visual tracking.
\newblock In {\em 2016 IEEE Conference on Computer Vision and Pattern
  Recognition (CVPR)}, pages 1430--1438, 2016.

\bibitem{2016Convolutional}
Martin Danelljan, Gustav Hager, Fahad~Shahbaz Khan, and Michael Felsberg.
\newblock Convolutional features for correlation filter based visual tracking.
\newblock In {\em 2015 IEEE International Conference on Computer Vision
  Workshop (ICCVW)}, 2016.

\bibitem{danelljan2017discriminative}
Martin {Danelljan}, Gustav {Hager}, Fahad~Shahbaz {Khan}, and Michael
  {Felsberg}.
\newblock Discriminative scale space tracking.
\newblock {\em IEEE Transactions on Pattern Analysis and Machine Intelligence},
  39(8):1561--1575, 2017.

\bibitem{2014Accurate}
Martin Danelljan, Gustav Häger, Fahad~Shahbaz Khan, and Michael Felsberg.
\newblock Accurate scale estimation for robust visual tracking.
\newblock In {\em British Machine Vision Conference}, 2014.

\bibitem{du2018the}
Dawei {Du}, Yuankai {Qi}, Hongyang {Yu}, Yifan {Yang}, Kaiwen {Duan}, Guorong
  {Li}, Weigang {Zhang}, Qingming {Huang}, and Qi {Tian}.
\newblock The unmanned aerial vehicle benchmark: Object detection and tracking.
\newblock In {\em Proceedings of the European Conference on Computer Vision
  (ECCV)}, pages 375--391, 2018.

\bibitem{duffner2013pixeltrack}
Stefan Duffner and Christophe Garcia.
\newblock Pixeltrack: a fast adaptive algorithm for tracking non-rigid objects.
\newblock In {\em Proceedings of the IEEE international conference on computer
  vision}, pages 2480--2487, 2013.

\bibitem{2019Correlation}
Changhong Fu, Fuling Lin, Yiming Li, and Guang Chen.
\newblock Correlation filter-based visual tracking for uav with online
  multi-feature learning.
\newblock {\em Remote Sensing}, 11(5), 2019.

\bibitem{2020Object}
Changhong Fu, Juntao Xu, Fuling Lin, Fuyu Guo, and Zhijun Zhang.
\newblock Object saliency-aware dual regularized correlation filter for
  real-time aerial tracking.
\newblock {\em IEEE Transactions on Geoscience and Remote Sensing}, page~12,
  2020.

\bibitem{2017Learning}
Hamed~Kiani Galoogahi, Ashton Fagg, and Simon Lucey.
\newblock Learning background-aware correlation filters for visual tracking.
\newblock In {\em 2017 IEEE International Conference on Computer Vision
  (ICCV)}, 2017.

\bibitem{godec2011hough}
Martin {Godec}, Peter~M. {Roth}, and Horst {Bischof}.
\newblock Hough-based tracking of non-rigid objects.
\newblock In {\em 2011 International Conference on Computer Vision}, pages
  81--88, 2011.

\bibitem{he2016deep}
Kaiming He, Xiangyu Zhang, Shaoqing Ren, and Jian Sun.
\newblock Deep residual learning for image recognition.
\newblock In {\em Proceedings of the IEEE conference on computer vision and
  pattern recognition}, pages 770--778, 2016.

\bibitem{he2020towards}
Yujie {He}, Changhong {Fu}, Fuling {Lin}, Yiming {Li}, and Peng {Lu}.
\newblock Towards robust visual tracking for unmanned aerial vehicle with
  tri-attentional correlation filters.
\newblock {\em arXiv preprint arXiv:2008.00528}, 2020.

\bibitem{2015High}
Joao~F. Henriques, Rui Caseiro, Pedro Martins, and Jorge Batista.
\newblock High-speed tracking with kernelized correlation filters.
\newblock {\em IEEE Transactions on Pattern Analysis \& Machine Intelligence},
  37(3):583--596, 2015.

\bibitem{huang2019learning}
Ziyuan {Huang}, Changhong {Fu}, Yiming {Li}, Fuling {Lin}, and Peng {Lu}.
\newblock Learning aberrance repressed correlation filters for real-time uav
  tracking.
\newblock In {\em 2019 IEEE/CVF International Conference on Computer Vision
  (ICCV)}, pages 2891--2900, 2019.

\bibitem{karaduman2019uav}
Mücahit {Karaduman}, Ahmet~Cevahir {Cinar}, and Haluk {Eren}.
\newblock Uav traffic patrolling via road detection and tracking in anonymous
  aerial video frames.
\newblock {\em Journal of Intelligent and Robotic Systems}, 95(2):675--690,
  2019.

\bibitem{kawaguchi2016deep}
Kenji {Kawaguchi}.
\newblock Deep learning without poor local minima.
\newblock In {\em Advances in Neural Information Processing Systems (NeurIPS)},
  volume~29, pages 586--594, 2016.

\bibitem{li2020training}
Fan {Li}, Changhong {Fu}, Fuling {Lin}, Yiming {Li}, and Peng {Lu}.
\newblock Training-set distillation for real-time uav object tracking.
\newblock In {\em 2020 IEEE International Conference on Robotics and Automation
  (ICRA)}, pages 9715--9721, 2020.

\bibitem{li2018learning}
Feng {Li}, Cheng {Tian}, Wangmeng {Zuo}, Lei {Zhang}, and Ming-Hsuan {Yang}.
\newblock Learning spatial-temporal regularized correlation filters for visual
  tracking.
\newblock In {\em 2018 IEEE/CVF Conference on Computer Vision and Pattern
  Recognition}, pages 4904--4913, 2018.

\bibitem{li2019gradnet}
Peixia {Li}, Boyu {Chen}, Wanli {Ouyang}, Dong {Wang}, Xiaoyun {Yang}, and
  Huchuan {Lu}.
\newblock Gradnet: Gradient-guided network for visual object tracking.
\newblock In {\em 2019 IEEE/CVF International Conference on Computer Vision
  (ICCV)}, pages 6162--6171, 2019.

\bibitem{li2017visual}
Siyi {Li} and Dit~Yan {Yeung}.
\newblock Visual object tracking for unmanned aerial vehicles: A benchmark and
  new motion models.
\newblock In {\em Proceedings of the 31st AAAI Conference on Artificial
  Intelligence}, pages 4140--4146, 2017.

\bibitem{li2020asymmetric}
Shui-wang Li, Qian-bo Jiang, Qi-jun Zhao, Li Lu, and Zi-liang Feng.
\newblock Asymmetric discriminative correlation filters for visual tracking.
\newblock {\em Frontiers of Information Technology \& Electronic Engineering},
  21(10):1467--1484, 2020.

\bibitem{li2019target}
Xin {Li}, Chao {Ma}, Baoyuan {Wu}, Zhenyu {He}, and Ming-Hsuan {Yang}.
\newblock Target-aware deep tracking.
\newblock In {\em 2019 IEEE/CVF Conference on Computer Vision and Pattern
  Recognition (CVPR)}, pages 1369--1378, 2019.

\bibitem{li2020autotrack}
Yiming {Li}, Changhong {Fu}, Fangqiang {Ding}, Ziyuan {Huang}, and Geng {Lu}.
\newblock Autotrack: Towards high-performance visual tracking for uav with
  automatic spatio-temporal regularization.
\newblock In {\em 2020 IEEE/CVF Conference on Computer Vision and Pattern
  Recognition (CVPR)}, pages 11923--11932, 2020.

\bibitem{li2019augmented}
Yiming {Li}, Changhong {Fu}, Fangqiang {Ding}, Ziyuan {Huang}, and Jia {Pan}.
\newblock Augmented memory for correlation filters in real-time uav tracking.
\newblock {\em arXiv preprint arXiv:1909.10989}, 2019.

\bibitem{li2020keyfilter}
Yiming {Li}, Changhong {Fu}, Ziyuan {Huang}, Yinqiang {Zhang}, and Jia {Pan}.
\newblock Keyfilter-aware real-time uav object tracking.
\newblock In {\em 2020 IEEE International Conference on Robotics and Automation
  (ICRA)}, pages 193--199, 2020.

\bibitem{li2014a}
Yang {Li} and Jianke {Zhu}.
\newblock A scale adaptive kernel correlation filter tracker with feature
  integration.
\newblock In {\em European Conference on Computer Vision}, pages 254--265,
  2014.

\bibitem{lin2020bicf}
Fuling {Lin}, Changhong {Fu}, Yujie {He}, Fuyu {Guo}, and Qian {Tang}.
\newblock Bicf: Learning bidirectional incongruity-aware correlation filter for
  efficient uav object tracking.
\newblock In {\em 2020 IEEE International Conference on Robotics and Automation
  (ICRA)}, pages 2365--2371, 2020.

\bibitem{2016Monocular}
Shanggang Lin, Matthew~A. Garratt, and Andrew~J. Lambert.
\newblock Monocular vision-based real-time target recognition and tracking for
  autonomously landing an uav in a cluttered shipboard environment.
\newblock {\em Autonomous Robots}, 41(4), 2016.

\bibitem{lukezic2020d3s}
Alan {Lukezic}, Jiri {Matas}, and Matej {Kristan}.
\newblock D3s – a discriminative single shot segmentation tracker.
\newblock In {\em 2020 IEEE/CVF Conference on Computer Vision and Pattern
  Recognition (CVPR)}, pages 7133--7142, 2020.

\bibitem{lukezic2017discriminative}
Alan {Lukezic}, Tomas {Vojir}, Luka~Cehovin {Zajc}, Jiri {Matas}, and Matej
  {Kristan}.
\newblock Discriminative correlation filter with channel and spatial
  reliability.
\newblock In {\em 2017 IEEE Conference on Computer Vision and Pattern
  Recognition (CVPR)}, pages 4847--4856, 2017.

\bibitem{ma2015hierarchical}
Chao {Ma}, Jia-Bin {Huang}, Xiaokang {Yang}, and Ming-Hsuan {Yang}.
\newblock Hierarchical convolutional features for visual tracking.
\newblock In {\em 2015 IEEE International Conference on Computer Vision
  (ICCV)}, pages 3074--3082, 2015.

\bibitem{2019Rotation}
Seyed~Mojtaba Marvasti-Zadeh, Hossein Ghanei-Yakhdan, and Shohreh Kasaei.
\newblock Rotation-aware discriminative scale space tracking.
\newblock In {\em 2019 27th Iranian Conference on Electrical Engineering
  (ICEE)}, 2019.

\bibitem{2021Efficient}
Seyed~Mojtaba Marvasti-Zadeh, Hossein Ghanei-Yakhdan, and Shohreh Kasaei.
\newblock Efficient scale estimation methods using lightweight deep
  convolutional neural networks for visual tracking.
\newblock {\em Neural Computing and Applications}, pages 1--16, 2021.

\bibitem{1988Mixture}
G.~J. Mclachlan and K.~E. Basford.
\newblock Mixture models: Inference and applications to clustering.
\newblock {\em Applied Statistics}, 38(2), 1988.

\bibitem{2016A}
Matthias Mueller, Neil Smith, and Bernard Ghanem.
\newblock A benchmark and simulator for uav tracking.
\newblock {\em Far East Journal of Mathematical Sciences}, 2(2):445--461, 2016.

\bibitem{2017Context}
Matthias Mueller, Neil Smith, and Bernard Ghanem.
\newblock Context-aware correlation filter tracking.
\newblock In {\em IEEE Conference on Computer Vision \& Pattern Recognition},
  2017.

\bibitem{mueller2017context}
Matthias {Mueller}, Neil {Smith}, and Bernard {Ghanem}.
\newblock Context-aware correlation filter tracking.
\newblock In {\em 2017 IEEE Conference on Computer Vision and Pattern
  Recognition (CVPR)}, pages 1387--1395, 2017.

\bibitem{nash2000multigrid}
Stephen~G Nash.
\newblock A multigrid approach to discretized optimization problems.
\newblock {\em Optimization Methods and Software}, 14(1-2):99--116, 2000.

\bibitem{2015Towards}
Miguel Olivares-Mendez, Changhong Fu, Philippe Ludivig, Tegawendé Bissyandé,
  Somasundar Kannan, Maciej Zurad, Arun Annaiyan, Holger Voos, and Pascual
  Campo.
\newblock Towards an autonomous vision-based unmanned aerial system against
  wildlife poachers.
\newblock {\em Sensors}, 2015.

\bibitem{orhan2018skip}
A.~Emin {Orhan} and Xaq {Pitkow}.
\newblock Skip connections eliminate singularities.
\newblock In {\em International Conference on Learning Representations}, 2018.

\bibitem{ren2007tracking}
Xiaofeng {Ren} and J. {Malik}.
\newblock Tracking as repeated figure/ground segmentation.
\newblock In {\em 2007 IEEE Conference on Computer Vision and Pattern
  Recognition}, pages 1--8, 2007.

\bibitem{2004GrabCut}
C. Rother.
\newblock Grabcut : Interactive foreground extraction using iterated graph cut.
\newblock {\em Acm Trans Graph}, 23, 2004.

\bibitem{1950Adjustment}
Jack Sherman and Winifred~J Morrison.
\newblock Adjustment of an inverse matrix corresponding to a change in one
  element of a given matrix.
\newblock {\em The Annals of Mathematical Statistics}, 21(1):124--127, 1950.

\bibitem{song2017crest}
Yibing Song, Chao Ma, Lijun Gong, Jiawei Zhang, Rynson~WH Lau, and Ming-Hsuan
  Yang.
\newblock Crest: Convolutional residual learning for visual tracking.
\newblock In {\em Proceedings of the IEEE International Conference on Computer
  Vision}, pages 2555--2564, 2017.

\bibitem{sun2020fast}
Mingjie {Sun}, Jimin {Xiao}, Eng~Gee {Lim}, Bingfeng {Zhang}, and Yao {Zhao}.
\newblock Fast template matching and update for video object tracking and
  segmentation.
\newblock In {\em 2020 IEEE/CVF Conference on Computer Vision and Pattern
  Recognition (CVPR)}, pages 10791--10799, 2020.

\bibitem{szeliski2006locally}
Richard Szeliski.
\newblock Locally adapted hierarchical basis preconditioning.
\newblock In {\em ACM SIGGRAPH 2006 Papers}, pages 1135--1143. 2006.

\bibitem{tsai2018learning}
Yi-Hsuan Tsai, Ming-Yu Liu, Deqing Sun, Ming-Hsuan Yang, and Jan Kautz.
\newblock Learning binary residual representations for domain-specific video
  streaming.
\newblock In {\em Proceedings of the AAAI Conference on Artificial
  Intelligence}, volume~32, 2018.

\bibitem{2017End}
Jack Valmadre, Luca Bertinetto, João~F. Henriques, Andrea Vedaldi, and Philip
  H.~S. Torr.
\newblock End-to-end representation learning for correlation filter based
  tracking.
\newblock 2017.

\bibitem{vedaldi2010vlfeat}
Andrea {Vedaldi} and Brian {Fulkerson}.
\newblock Vlfeat: an open and portable library of computer vision algorithms.
\newblock In {\em Proceedings of the 18th ACM international conference on
  Multimedia}, pages 1469--1472, 2010.

\bibitem{voigtlaender2020siam}
Paul {Voigtlaender}, Jonathon {Luiten}, Philip~H.S. {Torr}, and Bastian
  {Leibe}.
\newblock Siam r-cnn: Visual tracking by re-detection.
\newblock In {\em 2020 IEEE/CVF Conference on Computer Vision and Pattern
  Recognition (CVPR)}, pages 6578--6588, 2020.

\bibitem{2018Kernel}
Chen Wang, Le Zhang, Lihua Xie, and Junsong Yuan.
\newblock Kernel cross-correlator.
\newblock In {\em The Thirty-Second AAAI Conference on Artificial Intelligence
  (AAAI-18)}, 2018.

\bibitem{wang2019unsupervised}
Ning {Wang}, Yibing {Song}, Chao {Ma}, Wengang {Zhou}, Wei {Liu}, and Houqiang
  {Li}.
\newblock Unsupervised deep tracking.
\newblock In {\em 2019 IEEE/CVF Conference on Computer Vision and Pattern
  Recognition (CVPR)}, pages 1308--1317, 2019.

\bibitem{2018Multi}
Ning Wang, Wengang Zhou, Qi Tian, Richang Hong, and Houqiang Li.
\newblock Multi-cue correlation filters for robust visual tracking.
\newblock In {\em 2018 IEEE/CVF Conference on Computer Vision and Pattern
  Recognition (CVPR)}, 2018.

\bibitem{wang2018multi}
Ning {Wang}, Wengang {Zhou}, Qi {Tian}, Richang {Hong}, Meng {Wang}, and
  Houqiang {Li}.
\newblock Multi-cue correlation filters for robust visual tracking.
\newblock In {\em 2018 IEEE/CVF Conference on Computer Vision and Pattern
  Recognition}, pages 4844--4853, 2018.

\bibitem{wang2019fast}
Qiang {Wang}, Li {Zhang}, Luca {Bertinetto}, Weiming {Hu}, and Philip~H.S.
  {Torr}.
\newblock Fast online object tracking and segmentation: A unifying approach.
\newblock In {\em 2019 IEEE/CVF Conference on Computer Vision and Pattern
  Recognition (CVPR)}, pages 1328--1338, 2019.

\bibitem{2017A}
Z.~L Wang and B.~G Cai.
\newblock A comparison study of adaptive scale estimation in correlation
  filter-based visual tracking methods.
\newblock {\em Robotics \& Biomimetics}, 4(1):11, 2017.

\bibitem{wen2018visdrone}
Longyin {Wen}, Pengfei {Zhu}, Dawei {Du}, and et al.
\newblock Visdrone-sot2018: The vision meets drone single-object tracking
  challenge results.
\newblock In {\em Proceedings of the European Conference on Computer Vision
  (ECCV) Workshops}, pages 469--495, 2018.

\bibitem{2013Online}
Yi Wu, Jongwoo Lim, and Ming~Hsuan Yang.
\newblock Online object tracking: A benchmark.
\newblock In {\em Computer Vision \& Pattern Recognition}, 2013.

\bibitem{2017Aerial}
Chi Yuan, Zhixiang Liu, and Youmin Zhang.
\newblock Aerial images-based forest fire detection for firefighting using
  optical remote sensing techniques and unmanned aerial vehicles.
\newblock {\em Journal of Intelligent \& Robotic Systems}, 2017.

\bibitem{2017Action}
Sangdoo Yun, Jongwon Choi, Youngjoon Yoo, Kimin Yun, and Jin~Young Choi.
\newblock Action-decision networks for visual tracking with deep reinforcement
  learning.
\newblock In {\em IEEE Conference on Computer Vision \& Pattern Recognition},
  2017.

\bibitem{2017Multi}
Tianzhu Zhang, Changsheng Xu, and Ming~Hsuan Yang.
\newblock Multi-task correlation particle filter for robust object tracking.
\newblock In {\em 2017 IEEE Conference on Computer Vision and Pattern
  Recognition (CVPR)}, 2017.

\end{thebibliography}
}

\end{document}